\documentclass[journal]{IEEEtran}
\usepackage{algorithm}
\usepackage{algpseudocode}
\usepackage{graphicx}
\usepackage{multirow}
\usepackage{subfigure}
\usepackage{ifpdf}
\usepackage{diagbox}
\usepackage{bbm}
\usepackage{amssymb}
\usepackage{booktabs}
\usepackage{makecell}
\usepackage{colortbl}  
\usepackage{xcolor}
\usepackage{float}
\usepackage{acronym}
\usepackage{amsmath}

\begin{document}

\title{Text Growing on Leaf}

\author{
Chuang~Yang,
Mulin~Chen,
Yuan~Yuan,~\IEEEmembership{Senior Member,~IEEE,}
and~Qi~Wang,~\IEEEmembership{Senior Member,~IEEE}

\thanks{
Chuang~Yang is with the School of Computer Science, and with the School of Artificial Intelligence, OPtics and ElectroNics (iOPEN), Northwestern Polytechnical University, Xi'an 710072, Shaanxi, P. R. China.
}

\thanks{
Mulin~Chen, Yuan~Yuan, and Qi~Wang are with the School of Artificial Intelligence, OPtics and Electronics (iOPEN), Northwestern Polytechnical University, Xi'an 710072, P.R. China.
}

\thanks{
E-mail: cyang113@mail.nwpu.edu.cn, chenmulin@mail.nwpu.edu.cn, y.yuan.ieee@gmail.com, crabwq@gmail.com.
}

\thanks{
Qi Wang is the corresponding author.
}
}


\maketitle

\begin{abstract}
Irregular-shaped texts bring challenges to Scene Text Detection (STD). Although existing contour point sequence-based approaches achieve comparable performances, they fail to cover some highly curved ribbon-like text lines. It leads to limited text fitting ability and STD technique application. Considering the above problem, we combine text geometric characteristics and bionics to design a natural leaf vein-based text representation method (LVT). Concretely, it is found that leaf vein is a generally directed graph, which can easily cover various geometries. Inspired by it, we treat text contour as leaf margin and represent it through main, lateral, and thin veins. We further construct a detection framework based on LVT, namely LeafText. In the text reconstruction stage, LeafText simulates the leaf growth process to rebuild text contour. It grows main vein in Cartesian coordinates to locate text roughly at first. Then, lateral and thin veins are generated along the main vein growth direction in polar coordinates. They are responsible for generating coarse contour and refining it, respectively. Considering the deep dependency of lateral and thin veins on main vein, the Multi-Oriented Smoother (MOS) is proposed to enhance the robustness of main vein to ensure a reliable detection result. Additionally, we propose a global incentive loss to accelerate the predictions of lateral and thin veins. Ablation experiments demonstrate LVT is able to depict arbitrary-shaped texts precisely and verify the effectiveness of MOS and global incentive loss. Comparisons show that LeafText is superior to existing state-of-the-art (SOTA) methods on MSRA-TD500, CTW1500, Total-Text, and ICDAR2015 datasets.
\end{abstract}


\section{Introduction}
\label{Introduction}
\IEEEPARstart{R}{eading} scene text helps intelligent devices are able to accomplish many applications (such as unmanned systems, intelligent transport, express system, and so on), which has dramatically improved production efficiency and people's quality of life. Scene Text Detection (STD)~\cite{wan2021self} is the key technique for intelligent devices to simulate humans reading scene text, which has attracted a growing number of researchers and becomes a hot topic in computer vision. In the past decade, deep learning has greatly promoted the development of many computer technologies. It helps to extract strong expressive image features for many tasks (e.g. recognition, tracking, and regression). Benefiting from the advantages of deep learning, the performance of STD technique achieves excellent improvements in the aspect of the regular-shaped text detection~\cite{shi2017detecting,DBLP:conf/cvpr/LiuLYFLG18}. However, there are many irregular-shaped texts in real scenarios, which brings challenges to traditional approaches. To fit arbitrary-shaped text instances effectively, an increasing number of novel methods are proposed, which can be categorized into segmentation-based methods~\cite{DBLP:journals/tip/YangCXYW22,fu2022learning,cai2022arbitrarily} and regression-based methods~\cite{DBLP:conf/cvpr/FengYZL21,xu2020gliding,wang2019arbitrary} roughly. 

\begin{figure}
	\centering
	\includegraphics[width=.45\textwidth]{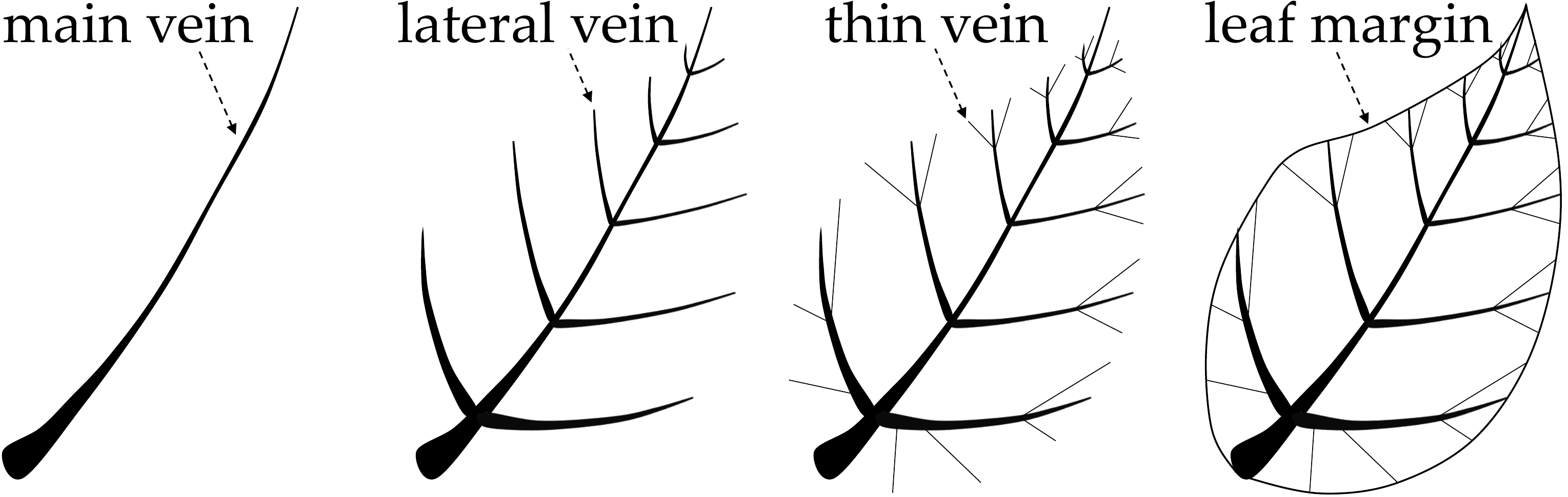}
	\caption{Illustration of the proposed leaf vein-based text representation method. We aim to treat text contour as leaf margin and construct it through main, lateral, and thin veins. Main vein is used for locating text instance roughly. Lateral vein is responsible for determining coarse contour. Accurate text contour is fined through thin vein.}
	\label{V1}
\end{figure}

The former adopts mask representation, which segments text regions directly and can detect irregular-shaped text instances naturally. However, these methods frequently require large training data and less supervision information aggravates this phenomenon. The latter represents text instances by contour point sequences. They try to sample point sequences by regressing the offsets between center point or quadrilateral and irregular-shaped contour, which have clear drawbacks. Specifically, the one-stage regression-based methods fail to fit highly curved ribbon-like text lines because multiple contour points may reside in the same direction. For multiple-stages methods, the intrinsically computationally expensive post-processing leads to low detection efficiency and limited practical applications. Therefore, how to design an efficient and effective text representation method is under explored.

Considering the limitations above, we combine text geometric characteristics and bionics to design a natural text representation method, which can fit text instances with any shapes accurately, even for highly curved ones. As shown in Figure~\ref{V1}, it is found that leaf margins always enjoy irregular shapes, which is similar to scene texts. Importantly, the leaf margin can be covered precisely by a directed graph that is composed of main, lateral, and thin veins. Inspired by the leaf vein structure, we propose to represent text contour by the combination of main, lateral, and thin veins. We further construct a one-stage text detection framework (called LeafText) based on leaf vein. It rebuilds text contours by simulating the leaf growth process, which is an elegant and effective design. Concretely, for one text instance, LeafText first grows the corresponding main vein from the predicted kernel mask in Cartesian coordinates to locate the text roughly. Then, lateral and thin veins sprout along both sides of the main vein growth direction in polar coordinates. In the end, the text contour is drawn by connecting endpoints of lateral and thin veins in a clockwise direction. Particularly, the lateral veins are used for determining coarse contour, and the thin veins are responsible for refining the contour to obtain an accurate detection result. Considering the deep dependencies of lateral and thin vein endpoints on the main vein, it is important to ensure a reliable main vein for rebuilding contour. However, the main vein extracted from the predicted kernel mask by the existing middle sampling method is always unreliable, which leads to a bad contour point sequence. Therefore, we propose a Multi-Oriented Smoother (MOS) to ensure the main vein robustness even when encountering unstable kernel masks. Additionally, text instances enjoy a large aspect ratio range compared with common objects, which brings challenges to predicting lateral and thin veins of text instances. Therefore, global incentive loss is proposed to force our model to balance the importance of text instances with different scales and focus on the prediction of lateral and thin veins. The main contributions of this paper are as follows:

\begin{enumerate}
	\item By combining the text geometric characteristics and bionics, a leaf vein-based text representation method (LVT) is proposed. It explores a natural and effective way to fit arbitrary-shaped text instances, which enhances the model's fitting ability.
	
	\item Thin vein is designed for fining text contours. It supports fitting texts accurately with lower model complexity, which accelerates the convergence of the training process effectively. Remarkably, the thin vein length is half of the lateral vein, which eases the learning of contour point sequence and ensures accurate detection results.
	
	\item A Multi-Oriented Smoother (MOS) is designed to ensure the robustness of the main vein extracted from the predicted kernel mask. It provides the correct growth directions to the lateral and thin veins, which ensures a reliable contour point sequence.
	
	\item Global incentive loss ${\cal L}_{g}$ is proposed to help balance the importance of text instances with different scales and force our method to focus on the predictions of lateral and thin veins. Particularly, it can be integrated into other regression-based detectors seamlessly.
\end{enumerate}

The rest of the paper is organized as follows. Section~\ref{Related Work} introduces the related works on text detection. Section~\ref{Methodology} describes the architecture, training process, and inference process details of LeafText. The experimental results are discussed in Section~\ref{Experiments}. Section~\ref{Conclusion} concludes the paper.

\section{Related Work}
\label{Related Work}
Recently, deep learning has promoted the development of the text detection technique greatly. According to the text representation method, previous text detectors can be classified into segmentation-based methods and regression-based methods roughly. In this section, a review of the existing text detection methods will be introduced.

\subsection{Segmentation-Based Methods}
Segmentation technology~\cite{long2015fully} executes pixel-level classification on images, which provides an effective solution for text detection. Zhang~$et~al$.~\cite{zhang2016multi} segmented rough text regions at first. Then, they extracted character components within text blocks by MSER~\cite{neumann2012real}. In the end, the authors suppressed false hypotheses by the intensity and geometric criteria of character components to obtain the final detection results. Lyu~$et~al$.~\cite{lyu2018multi} proposed to detect long text lines via a corner localization detection strategy. They generated candidate boxes by sampling and grouping corner points and filtered false-positive samples by the score of segmentation maps.

Deng~$et~al$.~\cite{DBLP:conf/aaai/DengLLC18} found that it would lead to text adhesion problems if extracting text contours from segmentation maps directly. To alleviate the above problem, link heat maps in eight directions were predicted for separating adhesive text instances. The works~\cite{tian2019learning,xu2019textfield} designed similar strategies as~\cite{DBLP:conf/aaai/DengLLC18} to provide solutions for the phenomenon that many texts are very close to each other. Different from the above works, Wang~$et~al$.~\cite{wang2019shape,wang2019efficient,wang2021pan++} and Liao~$et~al$.~\cite{liao2020real,liao2022real} proposed expansion strategies to generate text regions from shrink regions, which avoided detecting multiple adhesive texts as one either. The differences between them were that the former expanded shrink to text regions at pixel-level and the latter executed the expansion process at instance-level. The works~\cite{8812908,qin2019curved,dai2021accurate} considered that a small amount of pixel-level annotated data limits the model performance and proposed a two-stage detection framework to make full use of a large amount of data annotated with rectangles. In the inference process, the authors located texts roughly by quadrilaterals and extracted text contours precisely from the corresponding segmented text regions within quadrilaterals. Zhang~$et~al$.~\cite{zhang2020opmp} considered stack-omnidirectional text dilemma brings much challenges for text detection. They designed LSTM-based module to help generates omnidirectional text mask proposals from vertical and horizontal directions simultaneously to solve the stack-omnidirectional text dilemma.

\begin{figure*}
	\centering
	\includegraphics[width=.9\textwidth]{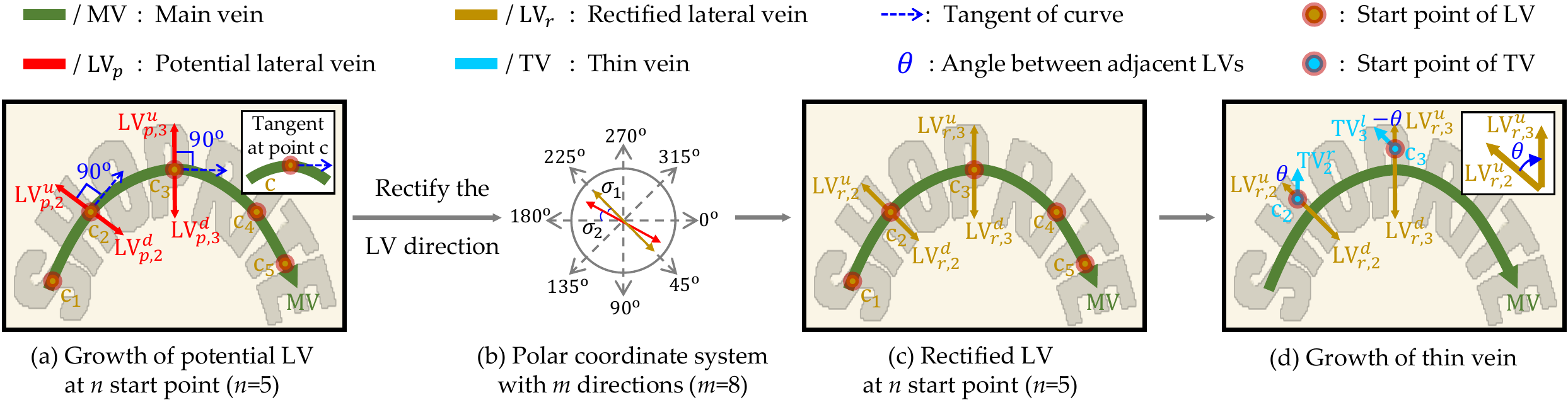}
	\caption{Illustration of the vein growth process. It contains the following three stages: 1) growing potential lateral vein, which is responsible for determining potential growth directions according to the start points and the corresponding tangent slopes; 2) rectifying potential growth lateral vein directions; 3) growing thin vein based on the determined lateral vein.}
	\label{V2}
\end{figure*}

Except for predicting the whole text instances directly, some approaches~\cite{baek2019character,DBLP:conf/cvpr/ZhangZHLYWY20,long2018textsnake} detect texts in character-level. Baek~$et~al$.~\cite{baek2019character} proposed a weakly-supervised framework to generate character-level labels to promote the training process. In the rebuilding process, the approach first predicted character regions and then linked them by affinities to obtain the final detection results. Zhang~$et~al$.~\cite{DBLP:conf/cvpr/ZhangZHLYWY20} adopted similar strategy as~\cite{baek2019character} to represent text instances. Moreover, they introduced Graph Convolutional Network (GCN) to predict the affinities between different character regions to improve the reliability of linked components. Different from them, Long~$et~al$.~\cite{long2018textsnake} segments center line firstly and then predict the each part bound along with the center line.

\subsection{Regression-Based Methods}
Object detection methods~\cite{ren2015faster,liu2016ssd,redmon2016you} adopt contour point sequence-based representation method to rebuild object contour or box, which brings great inspiration for the research of text detection. Liao~$et~al$.~\cite{liao2016textboxes} inherited the framework of~\cite{liu2016ssd} directly to detect horizontal text. To improve the performance of the multi-oriented texts detection, they proposed to predict rotation angles of texts in~\cite{liao2018textboxes++}. Different from the above anchor-based detection framework, Zhou~$et~al$.~\cite{zhou2017east} introduced the detection strategy proposed in~\cite{huang2015densebox} into text detection, which predicted corner points of multi-oriented texts and connected them to obtain the text boxes. Liao~$et~al$.~\cite{liao2018rotation} focused on how to extract strong expressive features for multi-oriented texts. They proposed to rotate the convolutional filters to encourage the model to extract rotation-sensitive features. He~$et~al.$~\cite{he2017single} extracted the text features with strong representation capacities through a hierarchical inception module. 

Though the above works achieve comparable performance in detecting multi-oriented text instances, they are hard to detect curved texts effectively. To improve the model's ability to detect arbitrary-shaped text instances, Some researchers~\cite{tang2019seglink++,hu2017wordsup,feng2019textdragon} separated word-level text blocks into multiple character-level regions. They regressed character boxes and linked those components to rebuild text blocks. The same as~\cite{tang2019seglink++}, Ma~$et~al.$~\cite{ma2021relatext} and Zhang~$et~al.$~\cite{zhang2020deep} adopted character-based detection strategy. Importantly, the authors utilized GCN to evaluate the linkages of adjacent characters to improve the stability of rebuilt text regions. Zhang~$et~al.$~\cite{zhang2019look} and Wang~$et~al.$~\cite{wang2020all} designed two-stage contour point sequence representation method. They extracted text quadrangles and further predicted contour points based on features within the quadrangles through regression way. The former generated the text center line (TCL) region at first. Then, they regressed the offset between TCL and text contour to sample the contour point. The latter predicted the distance between quadrangle and text contour directly to extract the contour point. Wang~$et~al.$~\cite{wang2020contournet} proposed a more intuitive way to obtain contour points. The authors segmented those points directly in both vertical and horizontal directions and combined them to filter unreliable results. Inspired by~\cite{xie2020polarmask}, Wang $et~al.$~\cite{wang2020textray} modeled text instances into the polar coordinate system and emitted multiple rays from text center to contour. The ray endpoints were sampled as contour points and connected to obtained final detection results. 

Some works~\cite{liu2020abcnet,su2022textdct,zhu2021fourier} proposed novel regression strategies to detect text instances and achieved state-of-the-art performance. Specifically, Liu~$et~al$.~\cite{liu2020abcnet} introduced Bezier-curve to represent text contours. They explored the probability to fit texts except for standard bounding box detection. Su~$et~al$.~\cite{su2022textdct} encode the text regions into compact vectors through discrete cosine transform. Zhu~$et~al$.~\cite{zhu2021fourier} modeled texts into Fourier domain and regressed contour point sequence by Fourier signature vectors. 

\section{Methodology}
\label{Methodology}
In this section, the leaf vein-based text representation method (LVT) is presented firstly. Then, we introduce the overall architecture of the proposed LeafText. Next, the details of Multi-Oriented Smoother (MOS) and Growth Process of Vein (GPV) are described. In the end, the optimization functions of network are given.

\begin{figure*}
	\centering
	\includegraphics[width=.85\textwidth]{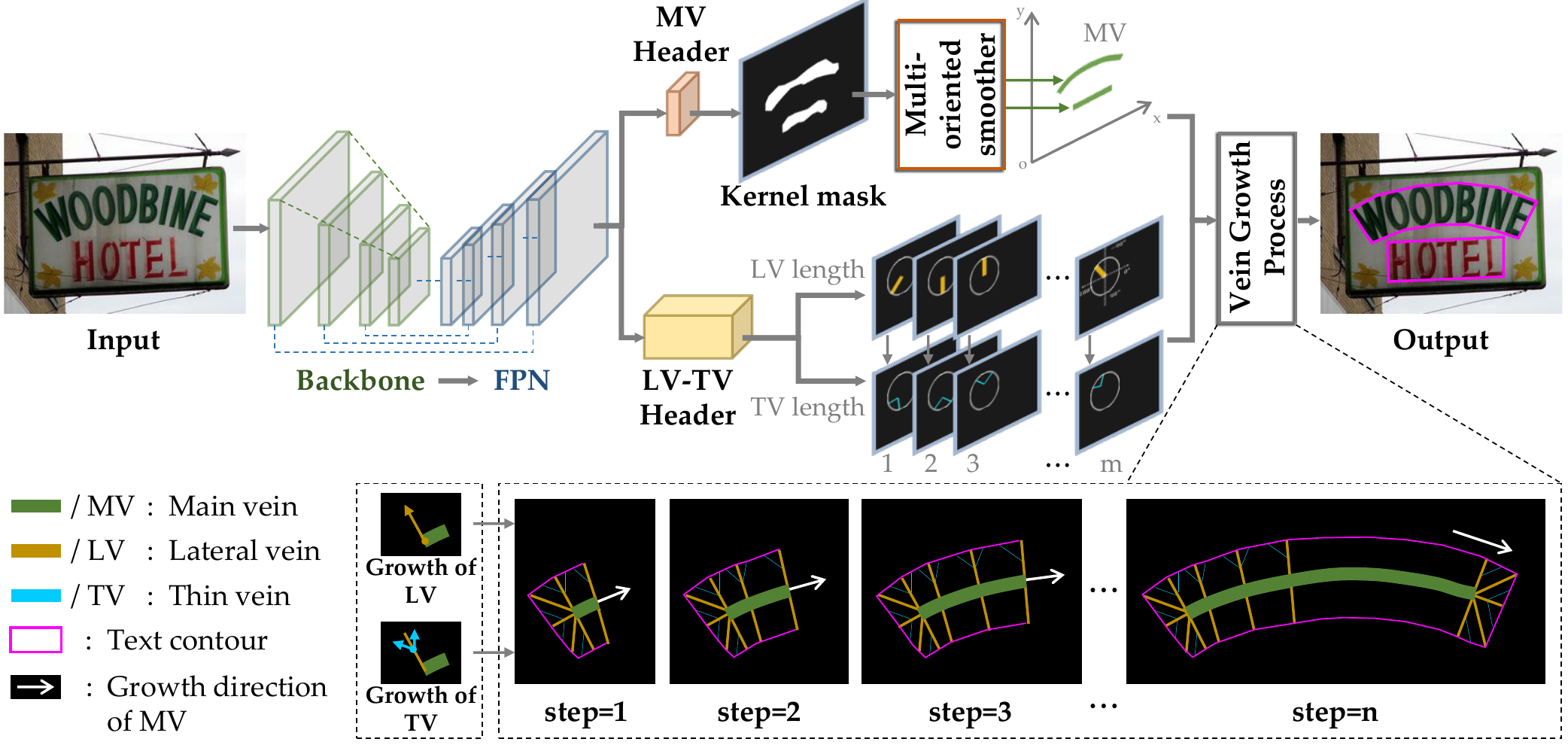}
	\caption{Overall architecture of the proposed LeafText. It is composed of Backbone, FPN, MV Header, LV-TV header, Multi-Oriented Smoother (MOS), and Vein Growth Process. MV, LV, and TV denote main vein, lateral vein, and thin vein respectively. MOS extracts main vein from kernel mask in Cartesian coordinate system. Vein Growth Process is the text contours reconstructing process in Fig.~\ref{V2}.}
	\label{V3}
\end{figure*}

\subsection{Leaf Vein-Based Text Representation Method}
\label{Leaf Vein-Based Text Representation Method}
The proposed text representation method (LVT) treats the text contour as leaf margin (as shown in Fig~\ref{V1}) and represents it through main, lateral, and thin veins, which can fit text instances with any shapes effectively even for highly curved ones. This section describes the growth process of veins mathematically combined with Fig~\ref{V2}.

For \textbf{main vein} (as we can see from Fig~\ref{V2}), it is used for locating texts roughly. The corresponding growth process is modeled as polynomial $f(x)$ by MOS (described in Section~\ref{Multi-Oriented Smoother}), which can be formulated as:
\begin{eqnarray}
\label{e1}
f\left( x \right) =\sum_{k=0}^K{\omega _kx^k}, \left( K\geqslant 1, x>0 \right) ,
\end{eqnarray}   
where $K$ is the degree of $f(x)$. $\omega_k$ is the coefficient of $x^k$.

Given a main vein $f(x)$, $n$ start points of lateral veins ($n$ is set to 5 for better visualization) are sampled equidistantly along the growth direction of main vein (sampling process can be referred to  Algorithm~\ref{algorithm1}). At each start point ($x_{lv},y_{lv}$) on $f(x)$, the corresponding tangent angle $\varphi_{lv}$ (Fig.~\ref{V2}~blue dotted arrow) can be computed by:
\begin{eqnarray}
\label{e2}
\varphi_{lv} =\mathrm{arc}\tan \left( \frac{df\left( x \right)}{dx}|_{x=x_{lv}} \right) .
\end{eqnarray}   

After obtaining $\varphi_{lv}$, we can determine the growth directions and lengths of lateral veins, which are responsible for generating coarse text contours.

For \textbf{the growth directions of lateral veins}, there are two lateral veins ($LV^{u}$ and $LV^{d}$) along the growth direction of main vein at each start point (as we can see from Fig.~\ref{V2}~(a)). The corresponding potential growth directions ($\mathrm{\alpha}_{p}^{u}$ and $\mathrm{\alpha}_{p}^{d}$) of them are defined as:
\begin{eqnarray}
\mathrm{\alpha}_{p}^{u}=\begin{cases}
\mathrm{\varphi}_{lv}-90^{\mathrm{o}},&		\mathrm{if}~ \mathrm{\varphi}_{lv}>90^{\mathrm{o}}\\
\mathrm{\varphi}_{lv}-90^{\mathrm{o}}+360^{\mathrm{o}},&		\mathrm{else}\\
\end{cases}, \\
\mathrm{\alpha}_{p}^{d}=\begin{cases}
\mathrm{\varphi}_{lv}+90^{\mathrm{o}},&		\mathrm{if}~ \mathrm{\varphi}_{lv}\leqslant 270^{\mathrm{o}}\\
\mathrm{\varphi}_{lv}+90^{\mathrm{o}}-360^{\mathrm{o}},&		\mathrm{else}\\
\end{cases} .
\end{eqnarray} 

Since the coverage of all 360 directions would lead to expensive computational costs, we predefined a polar coordinate system with $m$ directions ($m\ll 360$ and $m$ is set to 8 for better visualization) to rectify the potential direction of lateral vein, which avoids a highly complicated neural network and ensures the strong fitting ability to text contours (verified in Section~\ref{Upper Bound Analysis of LVT}). Concretely, as shown in Fig.~\ref{V2}~(b), supposing $\mathrm{\alpha}_{1}, \mathrm{\alpha}_{2}, ..., \mathrm{\alpha}_{M}$ are the all directions in the predefined polar coordinate system and $\mathrm{\alpha}_{m}\leqslant \mathrm{\alpha}_{p}^{u}\leqslant \mathrm{\alpha}_{m+1}\,\,\left( 1\leqslant {m}\leqslant M, m+1=1|{m=M} \right) $, the rectified direction $\mathrm{\alpha}_{rec}^{u}$ in Fig.~\ref{V2}~(c) can be calculated as: 
\begin{eqnarray}
\begin{gathered}
\label{e5}
\mathrm{\sigma}_1=|\mathrm{\alpha}_{m+1}-\mathrm{\alpha}_{p}^{u}|,
\\
\mathrm{\sigma}_2=|\mathrm{\alpha}_{p}^{u}-\mathrm{\alpha}_{m}|,
\\
\mathrm{\alpha}_{rec}^{u}=\begin{cases}
\mathrm{\alpha}_{m+1},&		\mathrm{if}~\mathrm{\sigma}_1<\mathrm{\sigma}_2\,\,\\
\mathrm{\alpha}_{m},&		\mathrm{else}\\
\end{cases},
\end{gathered}
\end{eqnarray} 
where $\mathrm{\sigma}_1$ and $\mathrm{\sigma}_2$ are the angles between the potential direction and the corresponding two adjacent directions in the predefined polar coordinate system. $|\cdot|$ denotes the operator for absolute value.

For \textbf{the lengths of lateral veins}, it can be constructed as the distances between the start points of lateral veins and text contours along the growth directions of lateral veins (refered to ~\ref{Label Generation}).

With the determined lateral veins, the growth directions and lengths of thin veins can be formulated. They are used for refining the coarse contours generated by lateral veins to reconstruct accurate detection results. As we can see from Fig.~\ref{V2}~(d), the middle points of lateral veins are sampled as the start points of thin veins, and there are two thin veins ($TV^{l}$ and $TV^{r}$) along the growth direction of lateral vein at one start point. For \textbf{the growth directions of thin veins}, they are determined according to the rectified directions of lateral veins. Specifically, given two adjacent lateral veins ($TV_{2}^{u}$ and $TV_{3}^{u}$) and the corresponding rectified growth directions ($\alpha_{rec,2}^{u}$ and $\alpha_{rec,3}^{u}$) that computed by Equation~\ref{e5}, we can obtain the growth directions ($\alpha_{2}^{r}$ and $\alpha_{3}^{l}$) of $TV_{2}^{r}$ and $TV_{3}^{l}$ by the Equation~\ref{e6} and Equation~\ref{e7}, respectively:
\begin{eqnarray}
\label{e6}
\alpha_{2}^{r}=\alpha_{rec,3}^{u} ,
\end{eqnarray} 
\begin{eqnarray}
\label{e7}
\alpha_{3}^{l}=\alpha_{rec,2}^{u} .
\end{eqnarray} 

For \textbf{the lengths of thin veins}, it is evaluated by the distances between the start points of thin veins and text contours along the growth directions of thin veins (refered to~\ref{Label Generation}). With the determined main veins, lateral veins, and thin veins, the text contour can be drawn by connected the endpoints of lateral veins and thin veins in a clockwise direction.

\subsection{Overall Pipeline}
\label{Overall Pipeline}
The overall pipeline of LeafText is shown in Figure~\ref{V3}, which consists of backbone, FPN, MV header, LV-TV header, Multi-Oriented Smoother(MOS), and Vein Growth Process. ResNet~\cite{he2016deep} is adopted as the \textbf{backbone} to help extract basic input image $\mathbf{I}\in \mathbb{R}^{h\times w\times 3}$ features, where $h, w$ are the height, width of input image with 3 channel. It outputs multiple coarse and fine feature maps $\mathbf{F}_{\frac{1}{s_1}}, \mathbf{F}_{\frac{1}{s_2}}, \mathbf{F}_{\frac{1}{s_3}}, \mathbf{F}_{\frac{1}{s_4}}$ simultaneously ($\mathbf{F}_{\frac{1}{s}}\in \mathbb{R}^{(h/s)\times (w/s)\times c}, s_1=4, s_2=8, s_3=16, s_4=32$), where $s$ denotes the stride of network and $c$ means feature maps channel. The coarse features bring a global correlation between texts and the fine ones focus on local details. To extract strong expressive features that are equipped with global and local information for the following detection headers, FPN~\cite{lin2017feature} is used for combining multiple features from the backbone to generate a concatenated feature map $\mathbf{F}_{concat}\in \mathbb{R}^{\left( h/4 \right) \times \left( w/4 \right) \times \left( c\times 4 \right)}$. As described in Section~\ref{Leaf Vein-Based Text Representation Method}, text contour is represented by the combination of main vein, lateral vein, and thin vein. To extract main vein of text instance, LeafText inputs the $\mathbf{F}_{concat}$ into MV header the generation of kernel mask map $\mathbf{F}_k\in \mathbb{R}^{\left( h/4 \right) \times \left( w/4 \right) \times 1}$ at first. Then, it extracts main vein from kernel mask $\mathbf{F}_k$ by MOS. For the growth of lateral and thin veins, LV-TV header conducts regression task on the $\mathbf{F}_{concat}$ to generate length mask map $\mathbf{F}_l\in \mathbb{R}^{\left( h/4 \right) \times \left( w/4 \right) \times m}$, where $m$ is the number of directions in predefined polar coordinate system (as described in Section~\ref{Leaf Vein-Based Text Representation Method}). In $\mathbf{F}_l$, pixel values at the start points of lateral and thin veins are the vein lengths in $m$ directions, respectively. With the determined main vein $f(x)$ (referred to Equation~\ref{e1}) and the lengths of lateral and thin veins in all $m$ directions, text contour can be generated by the Vein Growth Process (described to Section~\ref{Leaf Vein-Based Text Representation Method} and Fig.~\ref{V2}).

\begin{algorithm}  
	\label{algorithm1}
	\caption{Growth of Lateral Vein}  
	\begin{algorithmic}[1]
		\Require The kernel mask map $\mathbf{F}_{k}$;
		\Ensure  The coordinates of lateral vein start points $cpts^{e}_{r}$ and corresponding tangent slopes $\varphi^{e}$, len($cpts^{e}_{r}$)=len($\varphi^{e}$)=$n, n\geqslant 2$;
		
		\Function {Main}{$\mathbf{F}_{k}$}  
		\State $f(x)_{r} \gets$ \Call{Multi-Oriented Smoother}{$\mathbf{F}_{k}$}
		\State $cpts_{r}^{e} \gets $~equidistantSample($f(x)_{r}$) //$cpts_{r}^{e}$ means rotated equidistant start points of lateral veins sampled from $f(x)_{r}$, len($cpts_{r}^{e}$)=$n$;
		\State $\varphi_{r}^{e} \gets$~tangentSlope$(cpts_{r}^{e}, f(x)_{r})$ //$\varphi_{r}^{e}$ denote tangent slopes at start points, which can be computed by Equation~\ref{e2}, len($\varphi_{r}^{e}$)=$n$;
		\State $cpts^{e} \gets $rotate$(cpts_{r}^{e},-\phi,cpts[0])$
		\State $cpts^{e} \gets (cpts_{r}^{e}-h)$
		\State $\varphi^{e} \gets $rotate$(\varphi_{r}^{e},-\phi,cpts[0])$
		\State \Return{$cpts^{e},\varphi^{e}$}
		\EndFunction 
		\State
		
		\Function {Multi-Oriented Smoother}{$\mathbf{F}_{k}$}  
		\State $h~,w \gets $~size$(\mathbf{F}_{k})$  
		\State $kernel \gets $~padding$((\mathbf{F}_{k}>0),h,0)$ //obtaining kernel~mask $kernel$ from $\mathbf{F}_{k}$ and padding it by 0 in top, bottom, left, and right directions with $h$;
		\State $cpts \gets $ \Call{MiddleSample}{$kernel$} //$cpts$ denotes multiple center points of kernel mask, len($cpts$)=$n$;
		\State $\phi \gets $~angle$(\overrightarrow{cpts[0]cpts[-1]},\overrightarrow{Ox})$ //$\phi$ is the angle between vector $\overrightarrow{cpts[0]cpts[-1]}$ and vector $\overrightarrow{Ox}$ 
		\State $cpts_{r} \gets $~rotate$(cpts,\phi,cpts[0])$ // rotating the $cpts$ $\phi$ with $cpts[0]$ as the origin;
		\State $f(x)_{r} \gets $~polyFit$(cpts_{r})$ //$f(x)_{r}$ is rotated $f(x)$;
		\State \Return{$f(x)_{r}$}
		\EndFunction 
		\State
		
		\Function{MiddleSample}{$kernel$} 
		\State initial $cpts \gets [\oslash]$ 
		\State $pts_k \gets $coordinate$(kernel)$ //$pts_k$ are the point coordinates of kernel region;
		\State $x_{min} \gets $min$(pts^x_k)$,~~$x_{max} \gets $max$(pts^x_k)$ //$pts^x_k$ are the x coordinates of $pts_k$;
		\State $y_{min} \gets $min$(pts^y_k)$,~~$y_{max} \gets $max$(pts^y_k)$ // $pts^y_k$ are the y coordinates of $pts_k$;
		\For{$i = 1 \to n$}  
		\If{$x_{max}-x_{min}>y_{max}-y_{min}$}
		\State $x_{sample} \gets x_{min}+i\times \frac{x_{max}-x_{min}}{n+1}$  
		\State $y_{s} \gets pts_k[pts_k[:,0]==x_{sample}][:,1]$  
		\State $y_{s} \gets sort(y_{s})$  
		\State $y_{sample} \gets y_{s}[len(y_{s})//2]$  
		\Else
		\State $y_{sample} \gets y_{min}+i\times \frac{y_{max}-y_{min}}{n+1}$  
		\State $x_{s} \gets pts_k[pts_k[:,1]==y_{sample}][:,0]$  
		\State $x_{s} \gets sort(x_{s})$  
		\State $x_{sample} \gets x_{s}[len(x_{s})//2]$  
		\EndIf 
		\State $cpts[i] \gets (x_{sample},y_{sample})$ 
		\EndFor  
		\State \Return{$cpts$}
		\EndFunction  
	\end{algorithmic}  
\end{algorithm}

\subsection{Multi-Oriented Smoother}
\label{Multi-Oriented Smoother}
As shown in Fig.~\ref{V2} and Fig.~\ref{V3}, extracting main vein $f(x)$ (refer to Equation~\ref{e1}) from the predicted kernel mask accurately is important for determining the lateral and thin veins, which is the key to rebuild text contour. However, generating main vein by the existing middle sampling method always results in a discrete jagged result, which leads to bad growth directions for lateral and thin veins and unreliable reconstructed text contours.

Considering the above issue, Multi-Oriented Smoother (MOS) is designed to improve the reliability of main vein. Specifically, as shown in Algorithm~1~\textbf{function}~MULTI-ORIENTED SMOOTHER, MOS extracts kernel mask region from $\mathbf{F}_{k}$ at first. Meanwhile, considering rotating image leads to information loss of text instances at image borders, MOS pads kernel mask $kernel$ by 0 in top, bottom, left, and right directions with $h$ ($h$ is the height of input image). Then, initial center points $cpts$ are sampled by the \textbf{function}~MIDDLE SAMPLE in Algorithm~1. Next, the angle $\phi$ between $kernel$ and X-axis is computed and $cpts$ are rotated $\phi$ with $cpts[0]$ as origin. In the end, main vein $f(x)$ is fitted by the rotated center points $cpts_r$. With a smooth main vein, reliable start points and growth directions of lateral veins can be determined by \textbf{function}~MAIN in Algorithm~1, which improves the reliability of detection results significantly (verified in Section~\ref{Effectiveness of MOS}).

\begin{figure*}
	\centering
	\includegraphics[width=.9\textwidth]{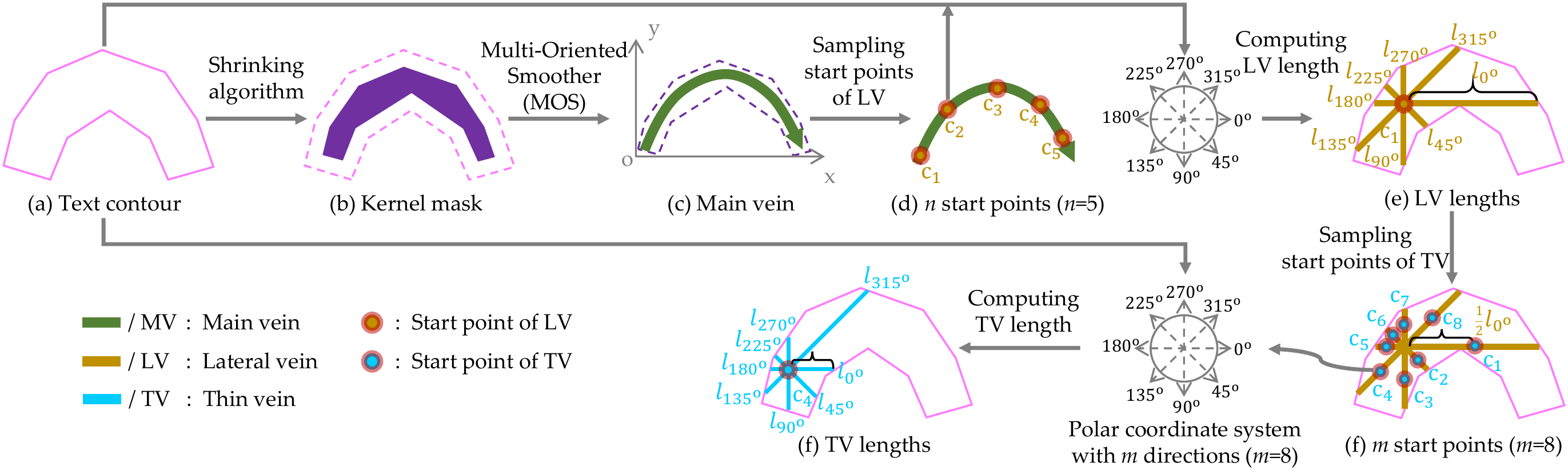}
	\caption{Visualization of the label generation process. Kernel mask (b) is used for extracting main vein. It is the ground-truth of MV header in Fig.~\ref{V3}. For Lengths of lateral and thin veins (e) and (f) in all directions of predefined polar coordinate system, they are responsible for supervising LV-TV header of LeafText in the training process.}
	\label{V4}
\end{figure*}

\subsection{Label Generation}
\label{Label Generation}
As described in Section~\ref{Leaf Vein-Based Text Representation Method}, text contour is represented by the combination of main, lateral, and thin veins. In Fig.~\ref{V3}, LeafText predicts kernel mask to extract main veins. Meanwhile, it regresses the lengths of lateral and thin veins in all directions of the predefined polar coordinate system. In this section, we illustrate the label generation process of kernel mask and vein lengths.

For \textbf{the label of kernel mask} (Fig.~\ref{V4}~(b)), the corresponding boundary is generated by shrinking text contour through the algorithm proposed in~\cite{vatti1992generic}. The inner region of the boundary is regarded as the kernel mask.

For \textbf{the label of lateral vein length}, main vein (Fig.~\ref{V4}~(c)) is extracted from kernel mask by the function~MULTI-ORIENTED SMOOTHER~in Algorithm~1 at first. Then, the start points and growth directions are determined (the details refer to function~MAIN~in Algorithm~1 and Section~\ref{Leaf Vein-Based Text Representation Method}, respectively). In the end, the lengths between start points and text contour in $m$ directions of the predefined polar coordinate system are computed. For \textbf{the label of thin vein length}, start points are sampled according to the lateral vein, and the lengths are computed in the same way as lateral vein.

\begin{figure*}
	\centering
	\includegraphics[width=.9\textwidth]{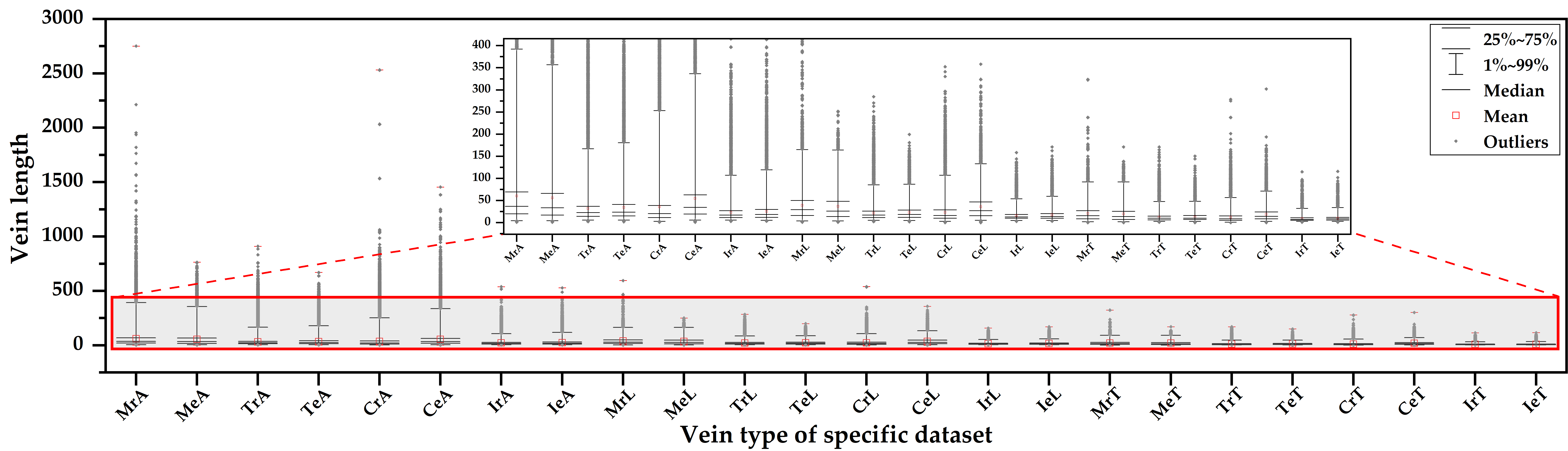}
	\caption{Illustration of the distributions of lateral and thin vein lengths on different public benchmarks. `M', `T', `C', and `I' indicate MSRA-TD500, Total-Text, CTW1500, and ICDAR2015 datasets respectively. `r' and `e'` are training and testing samples. `A' denotes the lengths in all directions (referred to Fig.~\ref{V4}). `L' and `T' are lengths in lateral and thin vein directions.}
	\label{V5}
\end{figure*}

\subsection{Loss Function}
\label{Loss Function}
LeafText determines main, lateral, and thin veins by MV and LV-TV Header (as shown in Fig.~\ref{V3}). In this paper, to optimize the proposed pipeline effectively, we propose a multi-task loss function ${\cal L}$ (referred to Equation~\ref{e8}). It consists of two loss functions of MV header loss ${\cal L}_{mv}$ and LV-TV header loss ${\cal L}_{lv-tv}$, which are responsible for supervising the corresponding headers in the training stage, respectively.
\begin{eqnarray}
\label{e8}
{\cal L}=\alpha {\cal L}_{mv}+\beta {\cal L}_{lv-tv} ,
\end{eqnarray}   
where $\alpha$ and $\beta$ are the coefficients of ${\cal L}_{mv}$ and ${\cal L}_{lv-tv}$. They are set to 1 and 0.25 in following experiments.

\textbf{Optimization of MV header.} Dice loss~\cite{milletari2016v} is designed for segmentation tasks where there is a strong imbalance between the positive and negative samples. Considering the regions of kernel mask are much smaller than background, Dice loss is adopted to evaluate the MV header loss ${\cal L}_{mv}$:
\begin{eqnarray}
{\cal L}_{mv} = 1-\frac{2\times| {\mathbf{K}_{pre}}\cap {\mathbf{K}_{gt}}|+\varepsilon}{| {\mathbf{K}_{pre}} |+| {\mathbf{K}_{gt}}|+\varepsilon},
\end{eqnarray}  
where $\mathbf{K}_{pre}$ and $\mathbf{K}_{gt}$ denote the predicted kernel mask and the corresponding ground-truth. To avoid the situation that there may be no positive samples in $\mathbf{K}_{gt}$, we set $\varepsilon$ as 1 to ensure that the denominator of ${\cal L}_{mv}$ is bigger than 0.

\textbf{Optimization of LV-TV header.} As we can see from Fig.~\ref{V3}, LV-TV header is responsible for regressing the lengths of lateral and thin veins. To facilitate the optimization of regression tasks, global incentive loss ${\cal L}_{g}$ is proposed to supervise LV-TV header in this paper:
\begin{eqnarray}
\begin{gathered}
{\cal L}_{lv-tv}={\cal L}_{g} .
\end{gathered}
\end{eqnarray}  

Global incentive loss ${\cal L}_{g}$ aims to force our model to balance the importance of text instances with different scales and focus on the prediction of lateral and thin veins.

Specifically, to keep the same sensitivity for multi-scale texts, ${\cal L}_{g}$ replaces L2-loss or Smooth-$l_1$ loss used in~\cite{redmon2016you,huang2015densebox} by negative logarithm loss ${\cal L}_{nl}$ (as shown in Equation~\ref{e11}). It scales the differences between predicted lengths and ground-truth into the range of 0--1, which ensures the effectiveness of our model to large and small text instances simultaneously.
\begin{eqnarray}
\label{e11}
{\cal L}_{nl}=-\log \left( \frac{\min (l_{pre},l_{gt})}{\max ( l_{pre},l_{gt})} \right) ,
\end{eqnarray}  
where $l_{pre}$ and $l_{gt}$ are the predicted lengths of lateral and thin veins and the corresponding ground-truth.

Moreover, as described in Section~\ref{Leaf Vein-Based Text Representation Method} and Section~\ref{Overall Pipeline}, LeafText determines one lateral vein or thin vein from $m$ directions. It leads to an overwhelming number of indirect samples and a small number of direct samples (lateral and thin veins). To make training more effective and efficient, we propose an incentive strategy for direct samples. It is found in Fig.~\ref{V5} that the lengths of direct samples are smaller than indirect ones for a specific dataset. Therefore, a 
incentive coefficient $\lambda$ is formulated as follow:
\begin{eqnarray}
\label{e12}
\lambda ={\rm tanh}(\rho (1-(l_{gt}/l_s))) ,
\end{eqnarray}  
where $l_s$ denotes the shorter side size of resized input images in the training and testing stages. $\rho$ is responsible for scaling $\lambda$ in to the range of 0--1.

By combining the Equation~\ref{e11} and~\ref{e12}, global incentive loss ${\cal L}_{g}$ can be formulated as:
\begin{eqnarray}
{\cal L}_{g} =\frac{1}{T\times M}\sum_{t=1}^T{\sum_{m=1}^M{\lambda ^{\left( t,m \right)}\mathcal{L}_{nl}^{\left( t,m \right)}}} ,
\end{eqnarray}  
where $T$ is the sum of number of all lateral and thin vein start points. $M$ denotes the direction number of the predefined polar coordinate system.

\section{Experiments}
\label{Experiments}
\subsection{Datasets}
To demonstrate the strong ability of LVT to fit arbitrary-shaped texts, we analyze the upper bound of the IoU between generated label and ground-truth. Meanwhile, the effectiveness of MOS and global incentive loss ${\cal L}_{g}$ are verified. Moreover, the proposed LeafText is evaluated on multiple representative public benchmarks to show the superior performance.

\textbf{SynthText}~\cite{gupta2016synthetic} contains 800k composite training samples that are combined by synthetic varied text instances and scene RGB images. It is proposed to pre-train the model to improve the robustness of the proposed LeafText.

\textbf{MSRA-TD500}~\cite{yao2012detecting} includes line-level Chinese and English text instances simultaneously. It is conposed of 300 training images and 200 testing images, respectively. To ensure a fair comparison environment, 400 images of HUST-TR400~\cite{yao2014unified} are extra introduced as training data.

\textbf{Total-Text}~\cite{ch2017total} consists of word-level arbitrary-shaped multilingual texts, which brings significant challenges for model generalization. There are 1255 images for training model and 300 images for evaluating performance.

\textbf{CTW1500}~\cite{yuliang2017detecting} is composed of 1500 samples, where includes 1000 training images and 500 testing images. Particularly, CTW1500 mainly contains line-level arbitrary-shaped text instances, which requires model's strong ability to deal with large-scale and ratio objects.

\textbf{ICDAR2015}~\cite{karatzas2015icdar} is proposed in ICDAR 2015 Robust Reading Competition, which has 1000 training image and 500 testing images. Different from the above three public benchmarks, the background of ICDAR2015 images are more complicated. Meanwhile, the text instances enjoys similar basic features with background, which brings much difficulties for text detection.

\begin{figure*}
	\centering
	\subfigure[Upper bound analysis of train dataset.]{
		\begin{minipage}[b]{0.9\linewidth}
			\includegraphics[width=1\linewidth]{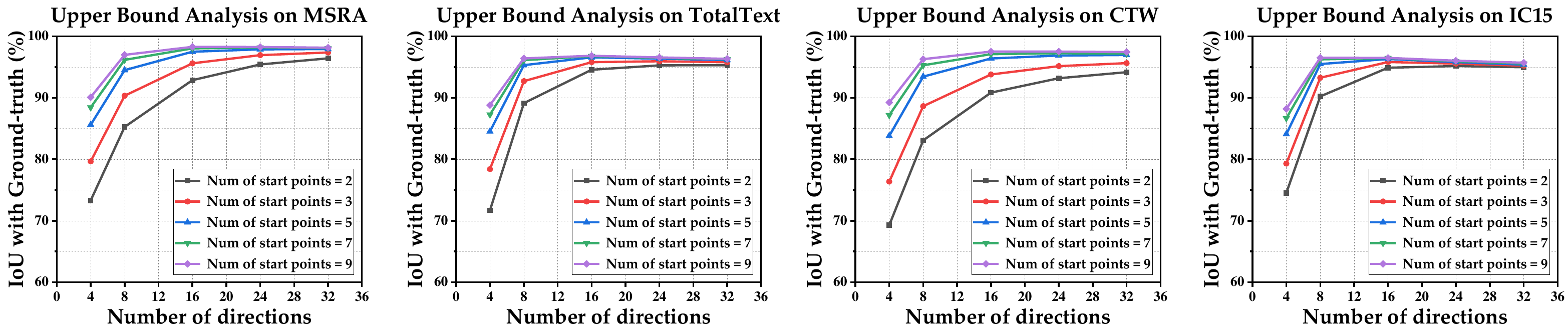}
	\end{minipage}}
	\subfigure[Upper bound analysis of test dataset.]{
		\begin{minipage}[b]{0.9\linewidth}
			\includegraphics[width=1\linewidth]{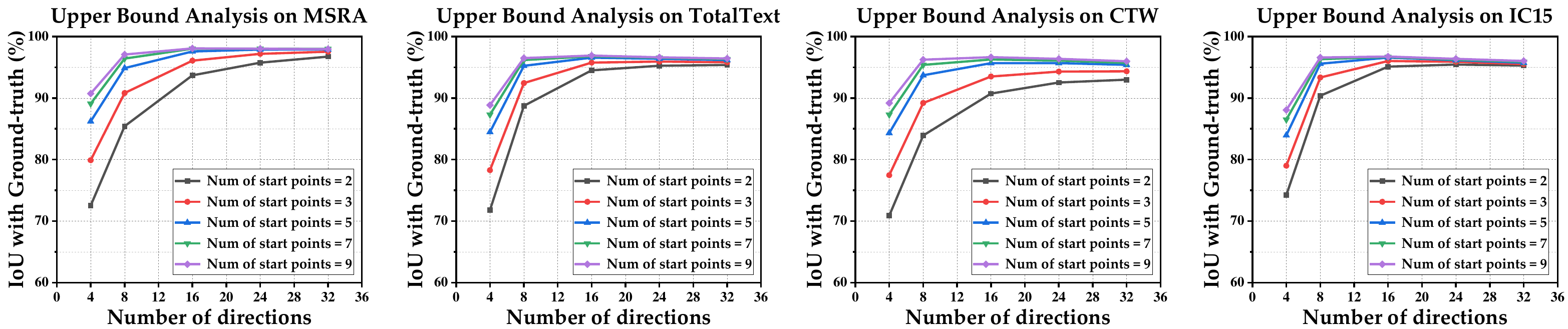}
	\end{minipage}}
	\caption{Upper Bound Analysis. More start points and directions of lateral vein can model text contours with higher IoU with Ground Truth. `Directions' are the vein directions in predefined polar coordinate system.}
	\label{V6}
\end{figure*}

\subsection{Implementation Details}
\label{Implementation Details}
The overall pipeline of the proposed LeafText is depicted in Fig.~\ref{V3}. The backbone adopts ResNet~\cite{he2016deep} directly and the details of FPN can be referred to~\cite{lin2017feature}. MV header and LV-TV header are composed of one $3\times 3$ convolutional layer and $m$ $3\times 3$ convolutional layer, respectively.

In the pre-process stage, training samples can be obtained through data augmentation and label generation operators. For the former, it contains the following strategies: (1) random scaling (including image size and aspect); (2) random horizontal flipping; (3) random rotating in the range of (-10°, 10°); (4) random cropping and padding. For the latter, kernel mask and the lengths of lateral and thin veins in $m$ directions are generated by the process in Fig.~\ref{V5}. Different from training samples, testing samples are produced by resizing input RGB images into specific sizes only. Particularly, the text instances labeled as DO NOT CARE are ignored during both the training and testing stages.

In the training stage, the weights of the CNN network are initialized first. Specifically, the backbone is pre-trained on the ImageNet~\cite{deng2009imagenet}. For the FPN and headers, they are initialized by the strategy proposed in~\cite{he2015delving}. To ensure an efficient and effective converge process, Adam~\cite{kingma2014adam} is adopted as the optimizer. The learning rate is set as 0.001 and adjusted through 'polylr' strategy with the model converging. In the comparison experiments, our model is trained on the SynthText dataset for 1 epoch at first. Then, it is finetuned on the official datasets (MSRA-TD500, Total-Text, CTW1500, and ICDAR2015) for 600 epochs with a batch size of 16. All the experiments in this paper are conducted on a workstation with RTX 1080Ti GPU.

\subsection{Ablation Study}
\label{Ablation Study}
To verify the effectiveness of LeafText, we conduct ablation experiments on multiple public benchmarks in this section. Specifically, to verify the strong fitting ability of LVT, we analyze the upper bound IoU between reconstructed text contour and ground-truth. Meanhiwle, the superiority of the proposed global incentive loss ${\cal L}_g$ is demonstrated by comparing it with exisiting loss functions. Furthermore, the importance of MOS for rebuilding text contours is verified. The details of experimental results are described in the following paragraphs.

\begin{figure*}
	\centering
	\includegraphics[width=.85\textwidth]{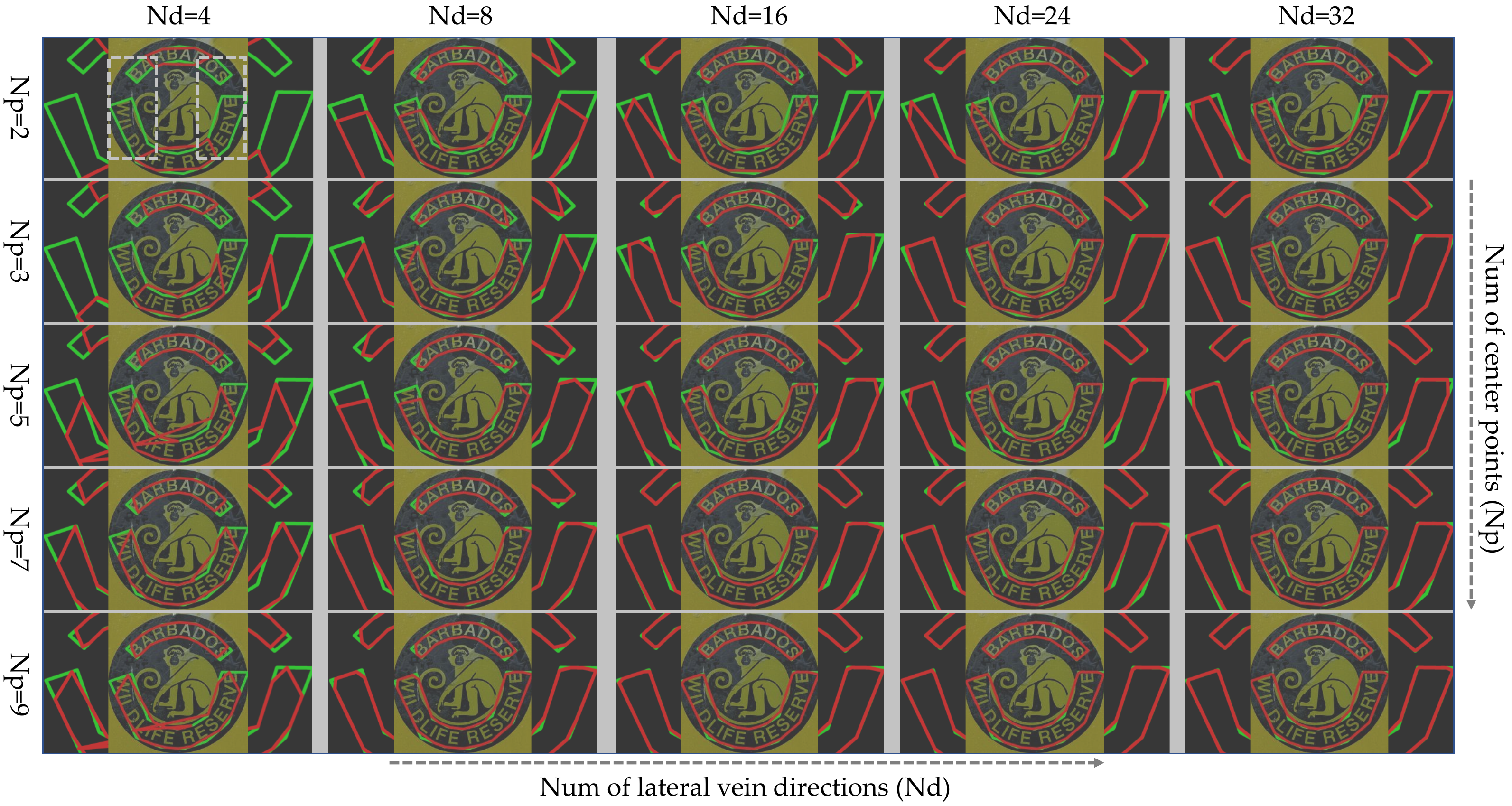}
	\caption{Visualization of the proposed leaf vein based text representation method. We aim to treat text contour as leaf margin and construct it through the combination of main vein, lateral vein, and thin vein.}
	\label{V7}
\end{figure*}
\begin{figure*}
	\centering
	\subfigure[Table of experimental results]{
		\begin{minipage}{0.4\linewidth}
			\renewcommand{\arraystretch}{1.3}
			\setlength{\tabcolsep}{2mm}
			\footnotesize
			\begin{tabular}{c|c||ccc||ccc}
				\Xhline{1pt}
				\multirow{2}{*}{$\rm N_d$} & \multirow{2}{*}{$\rm N_p$} & \multicolumn{3}{c||}{MSRA-TD500}                                          & \multicolumn{3}{c}{Total-Text}                                          \\ \cline{3-8} 
				&                     & \multicolumn{1}{c}{Precision} & \multicolumn{1}{c}{Recall} & F-measure & \multicolumn{1}{c}{Precision} & \multicolumn{1}{c}{Recall} & F-measure \\ 
				\Xhline{1pt}
				\multirow{5}{*}{4}  & 2                   & \multicolumn{1}{c}{86.9}      & \multicolumn{1}{c}{77.8}   & 82.1      & \multicolumn{1}{c}{78.6}      & \multicolumn{1}{c}{67.4}   & 72.6      \\ 
				& 3                   & \multicolumn{1}{c}{88.3}      & \multicolumn{1}{c}{78.9}   & 83.3      & \multicolumn{1}{c}{85.5}      & \multicolumn{1}{c}{73.4}   & 79.0      \\ 
				& 5                   & \multicolumn{1}{c}{88.5}      & \multicolumn{1}{c}{79.1}   & 83.5      & \multicolumn{1}{c}{89.0}      & \multicolumn{1}{c}{76.3}   & 82.1      \\ 
				& 7                   & \multicolumn{1}{c}{88.5}      & \multicolumn{1}{c}{79.1}   & 83.5      & \multicolumn{1}{c}{88.6}      & \multicolumn{1}{c}{75.9}   & 81.8      \\ 
				& 9                   & \multicolumn{1}{c}{88.2}      & \multicolumn{1}{c}{78.8}   & 83.2      & \multicolumn{1}{c}{89.1}      & \multicolumn{1}{c}{76.3}   & 82.2      \\ \hline\hline
				\multirow{5}{*}{8}  & 2                   & \multicolumn{1}{c}{90.7}      & \multicolumn{1}{c}{80.1}   & 85.1      & \multicolumn{1}{c}{88.8}      & \multicolumn{1}{c}{76.6}   & 82.3      \\ 
				& 3                   & \multicolumn{1}{c}{91.5}      & \multicolumn{1}{c}{80.4}   & 85.6      & \multicolumn{1}{c}{90.5}      & \multicolumn{1}{c}{78.2}   & 83.9      \\ 
				& 5                   &92.0&80.5&\textcolor{green}{\textbf{85.9}}& \multicolumn{1}{c}{91.1}      & \multicolumn{1}{c}{78.6}   & 84.4      \\ 
				& 7                   &91.9&80.4&\textcolor{blue}{\textbf{85.8}}& \multicolumn{1}{c}{91.2}      & \multicolumn{1}{c}{78.7}   & 84.5      \\ 
				& 9                   &92.1&80.6&\textcolor{red}{\textbf{86.0}}& \multicolumn{1}{c}{91.1}      & \multicolumn{1}{c}{78.6}   &\multicolumn{1}{c}{84.4}\\ \hline\hline
				\multirow{5}{*}{16} & 2                   & \multicolumn{1}{c}{88.7}      & \multicolumn{1}{c}{80.7}   & 84.5      & \multicolumn{1}{c}{88.6}      & \multicolumn{1}{c}{83.0}   & 85.7      \\ 
				& 3                   & \multicolumn{1}{c}{89.1}      & \multicolumn{1}{c}{81.0}   & 84.9      & 88.9&83.4&\textcolor{green}{\textbf{86.1}}      \\ 
				& 5                   & \multicolumn{1}{c}{89.1}      & \multicolumn{1}{c}{81.0}   & 84.9      & 89.0&83.5&\textcolor{red}{\textbf{86.2}}\\ 
				& 7                   & \multicolumn{1}{c}{88.9}      & \multicolumn{1}{c}{80.8}   & 84.7      &89.0&83.5&\textcolor{red}{\textbf{86.2}}\\ 
				& 9                   & \multicolumn{1}{c}{89.1}      & \multicolumn{1}{c}{81.0}   & 84.9      & 88.8&83.4&\textcolor{blue}{\textbf{86.0}}\\ \hline\hline
				\multirow{5}{*}{24} & 2                   & \multicolumn{1}{c}{86.6}      & \multicolumn{1}{c}{80.9}   & 83.7      & \multicolumn{1}{c}{89.3}      & \multicolumn{1}{c}{81.5}   & 85.2      \\ 
				& 3                   & \multicolumn{1}{c}{86.6}      & \multicolumn{1}{c}{80.9}   & 83.7      & \multicolumn{1}{c}{89.6}      & \multicolumn{1}{c}{81.8}   & 85.5      \\ 
				& 5                   & \multicolumn{1}{c}{87.2}      & \multicolumn{1}{c}{81.4}   & 84.2      & \multicolumn{1}{c}{89.3}      & \multicolumn{1}{c}{81.6}   & 85.3      \\ 
				& 7                   & \multicolumn{1}{c}{87.0}      & \multicolumn{1}{c}{81.2}   & 84.0      & \multicolumn{1}{c}{89.4}      & \multicolumn{1}{c}{81.7}   & 85.4      \\ 
				& 9                   & \multicolumn{1}{c}{87.2}      & \multicolumn{1}{c}{81.4}   & 84.2      & \multicolumn{1}{c}{89.4}      & \multicolumn{1}{c}{81.7}   & 85.4      \\ \hline\hline
				\multirow{5}{*}{32} & 2                   & \multicolumn{1}{c}{87.9}      & \multicolumn{1}{c}{79.8}   &  83.7         & \multicolumn{1}{c}{89.4}      & \multicolumn{1}{c}{76.2}   & 82.3      \\ 
				& 3                   & \multicolumn{1}{c}{88.5}      & \multicolumn{1}{c}{80.3}   & 84.2      & \multicolumn{1}{c}{89.6}      & \multicolumn{1}{c}{76.5}   & 82.5      \\ 
				& 5                   & \multicolumn{1}{c}{88.7}      & \multicolumn{1}{c}{80.5}   & 84.4      & \multicolumn{1}{c}{89.7}      & \multicolumn{1}{c}{77.4}   & 83.1      \\ 
				& 7                   & \multicolumn{1}{c}{88.7}      & \multicolumn{1}{c}{80.5}   & 84.4      & \multicolumn{1}{c}{89.1}      & \multicolumn{1}{c}{76.3}   & 82.2      \\ 
				& 9                   & \multicolumn{1}{c}{88.7}      & \multicolumn{1}{c}{80.5}   & 84.4      & \multicolumn{1}{c}{89.3}      & \multicolumn{1}{c}{76.4}   & 82.3      \\ 
				\Xhline{1pt}
			\end{tabular}
			\vspace{3mm}
	\end{minipage}}	
	\hspace{33mm}
	\subfigure[Curves of training loss and IoU.]{
		\begin{minipage}{0.332\linewidth}
			\includegraphics[width=1\linewidth]{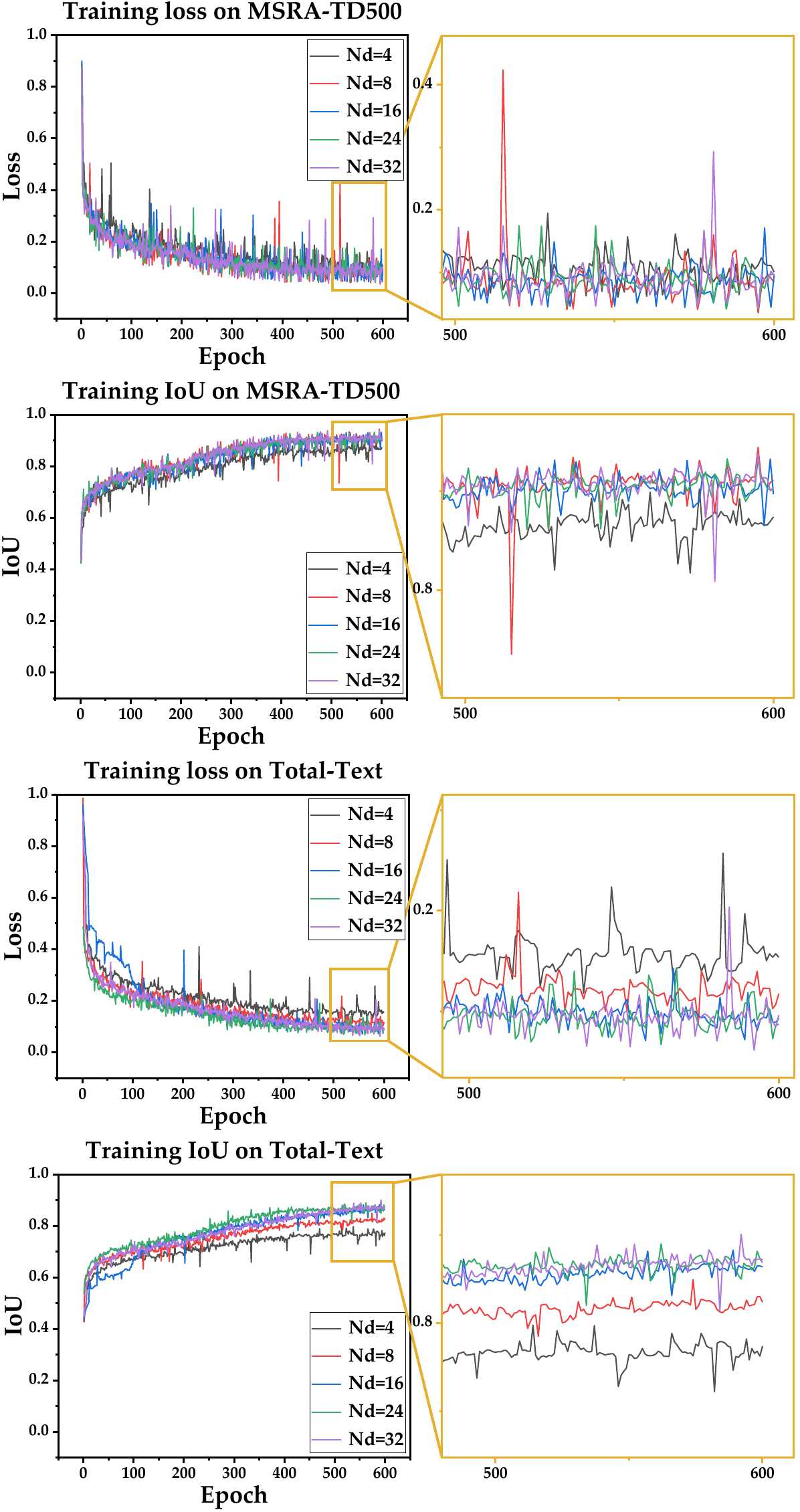}
			\vspace{.1mm}
	\end{minipage}}
	\caption{Ablation study for the impact of $\rm N_d$ and $\rm N_p$ on detection performance. $\rm N_d$ indicates the direction number predefined in a polar coordinate system. $\rm N_p$ means the number of center points sampled on the main vein for reconstructing text contours. \textcolor{red}{\textbf{red}}, \textcolor{green}{\textbf{green}}, and \textcolor{blue}{\textbf{blue}} are the experimental results with three best groups of settings respectively on MSRA-TD500 and Total-Text datasets. `IoU' in (b) indicates the Intersection of Union between predicted shrink-mask and the corresponding ground-truth.}
	\label{V8}
\end{figure*}

\textbf{Upper Bound Analysis of LVT.}
\label{Upper Bound Analysis of LVT}
Considering existing approaches fail to fit irregular-shaped texts accurately, a leaf vein-based text representation method is proposed. 

To verify the effectiveness of it, we analyze the upper bound of IoU that between rebuilt text contour based on generated label and ground-truth. Specifically, as shown in Fig.~\ref{V6}, the IoU can achieve 96\% at least on both training and testing samples of four public benchmarks (MSRA-TD500, Total-Text, CTW1500, and ICDAR2015). For some highly curved text instances (as visualized in Fig~\ref{V7}), LVT still can achieve superior performance. The results demonstrate the strong fitting ability of the proposed leaf vein-based text representation method for arbitrary-shaped texts.

Moreover, as described in Section~\ref{Leaf Vein-Based Text Representation Method}, the reconstruction process of LVT relies on the start points of lateral veins and the directions of predefined polar coordinate system. Therefore, we further explore the influences brought by the different numbers of the start points and directions (${\rm N_p}$ and ${\rm N_d}$ in Fig.~\ref{V7}). Concretely, as we can see from Fig.~\ref{V7}, the IoU is evaluated when tuning ${\rm N_p}$ and ${\rm N_d}$, respectively. It is found that there is a significant increase for IoU with ${\rm N_p}$ being tuned from 2 to 5. The upper bound of IoU continues to slow-growing when ${\rm N_p}$ is set to 7 and 9, which shows the start points of lateral veins play an important role in representing texts. Furthermore, the relations between IoU and ${\rm N_d}$ are visualized in this section. Concretely, in Fig.~\ref{V7}, larger ${\rm N_d}$ brings improvements to the fitting ability of our method, which verifies the importance of ${\rm N_d}$ for leaf vein-based text representation method.

\textbf{Performance Analysis under Different ${\rm N_d}$ and ${\rm N_p}$.} We have analyzed the upper bound IoU of the proposed LVT on different kinds of text instances in Section~\ref{Upper Bound Analysis of LVT}. To further verify the model's performance, LeafText is trained and evaluated under different ${\rm N_d}$ and ${\rm N_p}$ on MSRA-TD500 and Total-Text text benchmarks. 

Specifically, as we can see from the table~(a) in Fig.~\ref{V8}, for multi-oriented texts (MSRA-TD500), LeafText achieves the best performance when ${\rm N_d}$ and ${\rm N_p}$ are set to 8 and 9, respectively. Meanwhile, the F-measure begins to decrease with the increase of ${\rm N_d}$. For irregular-shaped text instances, our method achieves 86.2\% in F-measure when ${\rm N_d}$ and ${\rm N_p}$ are set as 16 and 5, which outperforms the rest of the other models. The above results show the best settings of ${\rm N_d}$ and ${\rm N_p}$ to detect multi-oriented and irregular-shaped texts. Furthermore, we visualize the details of the training process in Fig.~\ref{V8}~(b). It can be found from the curves of the training IoU on MSRA-TD500 that the IoU is smaller than the model under other settings when ${\rm N_d}$ equals 8, which matches the results of Table~(a) in Fig.~\ref{V8}. Meanwhile, the curves of the training IoU and loss on Total-Text show the unsatisfied convergence process when ${\rm N_d}$ is set to 4 and 8, which verifies the effectiveness of large ${\rm N_d}$ for irregular-shaped texts. The above experimental results provide appropriate model settings for the following comparison experiments on different kinds of text instances.

\begin{table}[]
	\renewcommand{\arraystretch}{1.3}
	\setlength{\tabcolsep}{3.0mm}
	\caption{The detection results of the models equipped with MOS and w/o MOS on MSRA-TD500 and Total-Text datasets.}
	\centering
	\begin{tabular}{c|cc|ccc}
		\Xhline{1pt}
		\multirow{2}{*}{MOS} & \multirow{2}{*}{$\rm N_d$} & \multirow{2}{*}{$\rm N_p$} & \multicolumn{3}{c}{MSRA-TD500}                                          \\ \cline{4-6} 
		&                     &                     & \multicolumn{1}{c}{Precision} & \multicolumn{1}{c}{Recall} & F-measure \\ \hline
		$\times$  &    \multirow{2}{*}{8}                 &       \multirow{2}{*}{9}              & \multicolumn{1}{c}{91.6}          & \multicolumn{1}{c}{78.8}       &     84.7      \\ 
		\checkmark&                     &                     & \multicolumn{1}{c}{92.1}          & \multicolumn{1}{c}{80.6}       &     86.0      \\ \Xhline{1pt}
		\multirow{2}{*}{MOS} & \multirow{2}{*}{$\rm N_d$} & \multirow{2}{*}{$\rm N_p$} & \multicolumn{3}{c}{Total-Text}                                          \\ \cline{4-6} 
		&                     &                     & \multicolumn{1}{c}{Precision} & \multicolumn{1}{c}{Recall} & F-measure \\ \hline
		$\times$ &    \multirow{2}{*}{16}                 &      \multirow{2}{*}{5}               & \multicolumn{1}{c}{88.3}          & \multicolumn{1}{c}{82.1}       &      85.1     \\ 
		\checkmark&                     &                     & \multicolumn{1}{c}{89.0}          & \multicolumn{1}{c}{83.5}       &      86.2     \\ \Xhline{1pt}
	\end{tabular}
	\label{table1}
\end{table}

\begin{figure*}
	\centering
	\includegraphics[width=.95\textwidth]{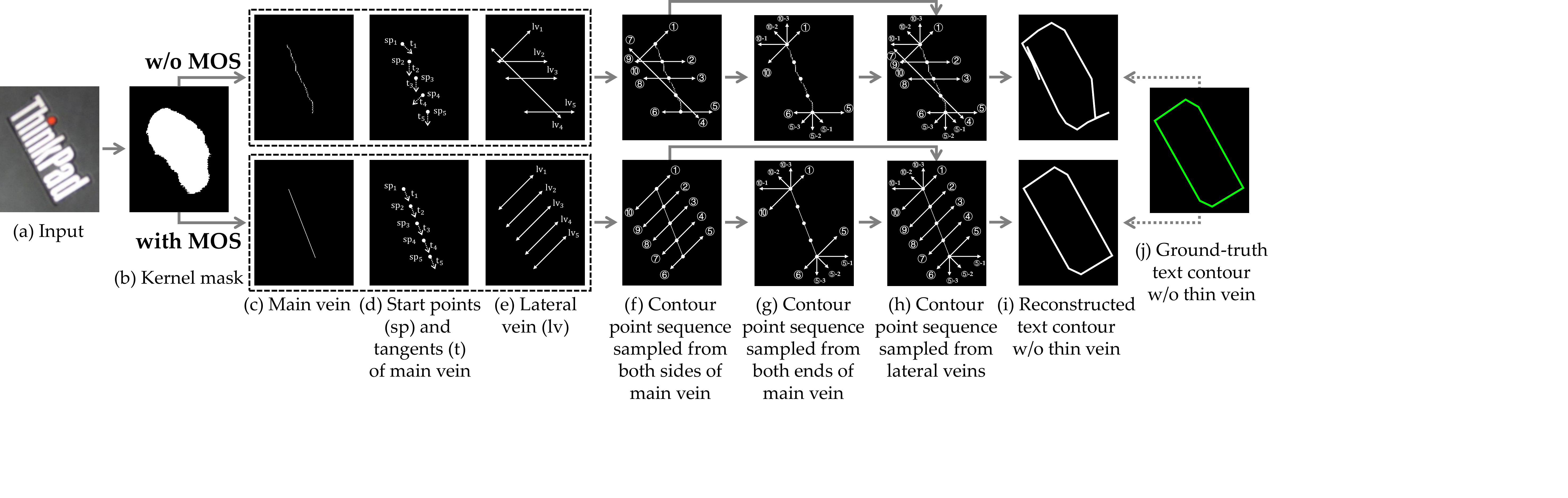}
	\caption{Visualization of the differences between the text contour reconstruction processes with MOS and w/o MOS. The sample is picked from MSRA-TD500 dataset and the $\rm N_d$ and $\rm N_p$ of model are set to 8 and 5, respectively.}
	\label{V9}
\end{figure*}

\textbf{Effectiveness of MOS.}
\label{Effectiveness of MOS}
As described in Section~\ref{Multi-Oriented Smoother}, for improving the accuracy of reconstructed text instance, MOS is designed to ensure the reliability of the main vein extracted from the predicted unreliable kernel mask. To demonstrate the effectiveness of MOS, we analyze the improvements in detection performance brought by MOS and visualize some qualitative results. As shown in Table~\ref{table1}, MOS can bring improvements in F-measure on both multi-oriented (MSRA-TD500) and irregular-shaped (Total-Text) datasets. Specifically, LeafText with MOS achieves 86.0\% and 86.2\% F-measure on the two benchmarks respectively, which surpasses LeafText without MOS 1.3\% and 1.1\%. These experimental results demonstrate the effectiveness of MOS for improving the quality of rebuilt contours. To further explain how MOS works for smoothing the main vein, we visualize the process details in Fig.~\ref{V9}. Concretely, given a predicted kernel mask (Fig.~\ref{V9}~(b)), MOS helps our method to determine correct tangent directions on each start point (Fig.~\ref{V9}~(d)), which helps avoid disordered contour point sequence (Fig.~\ref{V9}~(f)) and improve the reliability of reconstructed text contour (Fig.~\ref{V9}~(i)) effectively. The visualization demonstrates the effectiveness of MOS and depicts the differences between the text contour reconstruction processes with MOS and w/o MOS vividly.

\begin{table}[]
	\renewcommand{\arraystretch}{1.3}
	\setlength{\tabcolsep}{1.5mm}
	\caption{The detection results of the models trained by different loss functions on MSRA-TD500 and Total-Text datasets.}
	\centering
	\begin{tabular}{c|c|ccc}
		\Xhline{1pt}
		Dataset                     & Loss                  & Precision & Recall & F-measure \\ \Xhline{1pt}
		\multirow{3}{*}{MSRA-TD500} & Smooth-L1             &   89.2        &    77.3    &     82.8      \\ 
		& L2                    &    90.0       &    71.1    &    79.4       \\ 
		& Global incentive &   92.1        &    80.6    &    86.0       \\ \hline
		\multirow{3}{*}{Total-Text} & Smooth-L1             &   88.5        &  80.1      &    84.1       \\ 
		& L2                    &     88.7      &    74.5    &    81.0       \\ 
		& Global incentive &     89.0      &    83.5    &    86.2       \\ \Xhline{1pt}
	\end{tabular}
	\label{table2}
\end{table}

\textbf{Effectiveness of Global Incentive Loss.} Considering existing L2-loss and Smooth-$l_1$ loss mainly focus on large samples, which leads to the ignorance of small objects, global incentive loss ${\cal L}_{g}$ is designed to force our model to balance the importance of texts with different scales and focus on the prediction of lateral and thin veins.

\begin{figure}
	\centering
	\includegraphics[width=.42\textwidth]{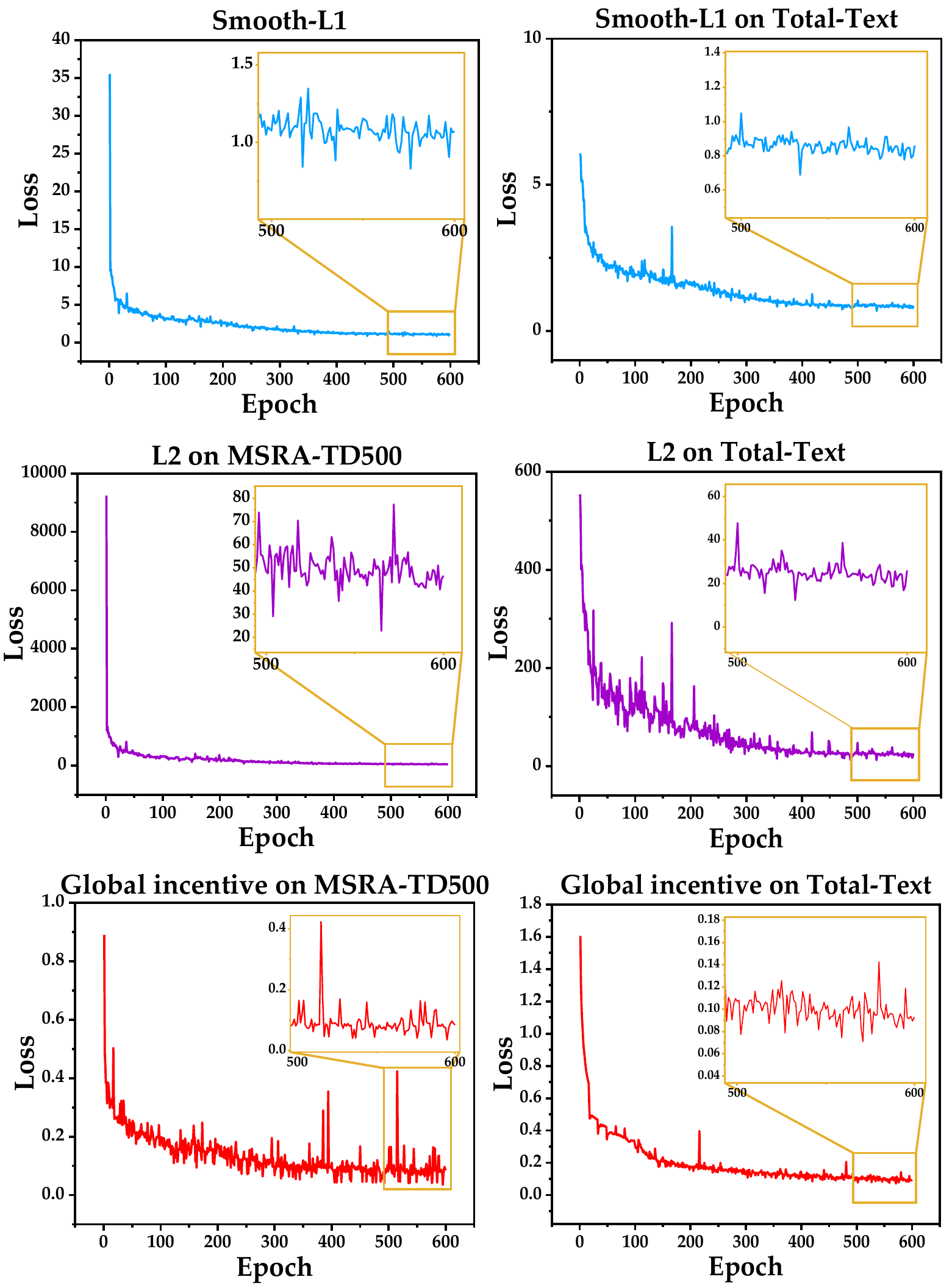}
	\caption{Visualization of the training processes on the MSRA-TD500 and Total-Text with different loss functions.}
	\label{V10}
\end{figure}

As shown in Table~\ref{table2}, compared with existing L2-loss and Smooth-$l_1$ loss, training LeafText by the proposed ${\cal L}_{g}$ brings 3.2\% and 2.1\% improvements in F-measure on MSRA-TD500 and Total-Text at least, respectively. Considering there existing lots of large and small texts simultaneously in MSRA-TD500, the above experimental results demonstrate ${\cal L}_{g}$ can help the model improve the ability to deal with different sized text instances. Meanwhile, the results in Table~\ref{table2} verify that our method can regress lateral and thin vein lengths more accurately when supervising the LV-TV prediction header by ${\cal L}_{g}$. Furthermore, we visualize the training processes of different losses in Fig.~\ref{V10}. It is found that global incentive loss function ${\cal L}_{g}$ fluctuates around 0.1 at the end of the convergence process on MSRA-TD500 and Total-Text datasets simultaneously. Compared with the loss functions of Smooth-L1 and L2, the proposed ${\cal L}_{g}$ can accelerate the model converging effectively and improve the model's ability to learn text features. The above results demonstrate the effectiveness of the proposed global incentive loss ${\cal L}_{g}$ for detecting multi-scaled texts.

\begin{table}[]
	\renewcommand{\arraystretch}{1.3}
	\setlength{\tabcolsep}{3.3mm}
	\caption{Impact of TV for detection results on  MSRA-TD500 and Total-Text datasets. `LV' and `TV' denote lateral vein and thin vein, respectively. `MAE' means Mean Absolute Error.}
	\centering
	\begin{tabular}{c|cc|ccc}
		\Xhline{1pt}
		\multirow{2}{*}{TV} & \multicolumn{2}{c|}{MAE}                                 & \multicolumn{3}{c}{MSRA-TD500}                                          \\ \cline{2-6} 
		& \multicolumn{1}{c}{LV}      & TV         & \multicolumn{1}{c}{Precision} & \multicolumn{1}{c}{Recall} & F-measure \\ \hline
		$\times$& \multicolumn{1}{c}{\multirow{2}{*}{11.3}} & \multirow{2}{*}{11.1} & \multicolumn{1}{c}{91.8}          & \multicolumn{1}{c}{80.2}       &   85.6         \\ 
		\checkmark& \multicolumn{1}{c}{}                  &                   & \multicolumn{1}{c}{92.1}          & \multicolumn{1}{c}{80.6}       &    86.0       \\ \Xhline{1pt}
		\multirow{2}{*}{TV} & \multicolumn{2}{c|}{MAE}                                 & \multicolumn{3}{c}{Total-Text}                                          \\ \cline{2-6} 
		& \multicolumn{1}{c}{LV}      & TV         & \multicolumn{1}{c}{Precision} & \multicolumn{1}{c}{Recall} & F-measure \\ \hline
		$\times$& \multicolumn{1}{c}{\multirow{2}{*}{5.8}} & \multirow{2}{*}{5.3} & \multicolumn{1}{c}{88.1}          & \multicolumn{1}{c}{82.7}       &   85.3       \\ 
		\checkmark& \multicolumn{1}{c}{}                  &                   & \multicolumn{1}{c}{89.0}          & \multicolumn{1}{c}{83.5}       &    86.2       \\
		\Xhline{1pt}
	\end{tabular}
	\label{table3}
\end{table}

\textbf{Superiority of the Thin Vein.} As described in Section~\ref{Leaf Vein-Based Text Representation Method}, the thin vein is designed for fining text contours, which supports accurately fitting texts with lower model complexity. Benefiting from the advantage that the thin vein length is half of the lateral vein, the thin vein eases the learning of contour point sequence and ensures accurate detection results. To verify the superiority of the thin vein, we evaluate the accuracy of the lateral and thin vein in table~\ref{table3}. We first evaluate the Mean Absolute Error (MAE) of the lateral vein and the thin vein. It is found that the MAE of the lateral vein surpasses the thin vein 0.2 and 0.5 on MSRA-TD500 and Total-Text. It demonstrates the task of thin vein prediction is more accessible than the prediction of the lateral vein, which verifies the advantage of thin vein that can ease the learning of contour point sequence. Meanwhile, thin vein brings 0.4\% and 0.9\% in F-measure on MSRA-TD500 and Total-Text, respectively. The above experimental results prove thin vein can promote the model performance in the detection of text instances effectively. 

\subsection{Comparison with State-of-the-Art Methods}
\label{Comparison}
To demonstrate the superior performance of LeafText for detecting texts with arbitrary shapes, multi scales, and multilingual, we compare it with the existing state-of-the-art (SOTA) approaches on four representative public benchmarks (MSRA-TD500, Total-Text, CTW1500, and ICDAR2015) in this section. Meanwhile, the advantages of our method over previous methods are analyzed based on the comparisons and quality detection results.

\begin{table}[]
	\renewcommand{\arraystretch}{1.3}
	\setlength{\tabcolsep}{2.mm}
	\caption{Performance comparison on MSRA-TD500 dataset.}
	\centering
	\begin{tabular}{r|ccc}
		\Xhline{1pt}
		\multicolumn{1}{c|}{Methods} &  Precision  &   Recall  &   F-measure \\ 
		\Xhline{1pt}
		MOTD~\cite{zhang2016multi} (CVPR 2016)                & 83.0 & 67.0 & 74.0 \\ 
		EAST~\cite{zhou2017east} (CVPR 2017)                  & 87.3 & 67.4 & 76.1 \\ 
		SegLink~\cite{shi2017detecting} (CVPR 2017)           & 86.0 & 70.0 & 77.0 \\ 
		PixelLink~\cite{DBLP:conf/aaai/DengLLC18} (AAAI 2018) & 83.0 & 73.2 & 77.8 \\ 
		TextSnake~\cite{long2018textsnake} (ECCV 2018)        & 83.2 & 73.9 & 78.3 \\
		RRD~\cite{liao2018rotation} (CVPR 2018)               & 87.0 & 73.0 & 79.0 \\ 
		CornerNet~\cite{lyu2018multi} (CVPR 2018)             & 87.6 & 76.2 & 81.5 \\
		CRAFT~\cite{baek2019character} (CVPR 2019)            & 88.2 & 78.2 & 82.9 \\ 
		TextField~\cite{xu2019textfield} (TIP 2019)           & 87.4 & 75.9 & 81.3 \\ 
		SAE~\cite{tian2019learning} (CVPR 2019)               & 84.2 & 81.7 & 82.9 \\ 
		ATRR~\cite{wang2019arbitrary} (CVPR 2019)             & 85.2 & 82.1 & 83.6 \\ 
		PAN~\cite{wang2019efficient} (ICCV 2019)              & 84.4 & 83.8 & 84.1 \\ 
		DB~\cite{liao2020real} (AAAI 2020)                    & 90.4 & 76.3 & 82.8 \\ 
		DRRG~\cite{DBLP:conf/cvpr/ZhangZHLYWY20} (CVPR 2020)  & 88.1 & 82.3 & 85.1 \\ 
		OPMP~\cite{zhang2020opmp} (TMM 2021)                  & 86.0 & 83.4 & 84.7 \\ 
		PAN++~\cite{wang2021pan++} (TPAMI 2021)               & 85.3 & 84.0 & 84.7 \\ 
		SAVTD~\cite{DBLP:conf/cvpr/FengYZL21} (CVPR 2021)     & 89.2 & 81.5 & 85.2 \\ 
		GV~\cite{xu2020gliding} (TPAMI 2021)                  & 88.8 & 84.3 & 86.5 \\ 
		ReLaText~\cite{ma2021relatext} (PR 2021)              & 90.5 & 83.2 & \textcolor{green}{\textbf{86.7}} \\ 
		LPAP~\cite{fu2022learning} (TOMM 2022)                & 87.9 & 77.7 & 82.5 \\ 
		DC~\cite{cai2022arbitrarily} (PR 2022)                & 87.9 & 83.1 & 85.4 \\  
		Res18-DB++~\cite{liao2022real} (TPAMI 2022)           & 87.9 & 82.5 & 85.1 \\
		Res50-DB++~\cite{liao2022real} (TPAMI 2022)           & 91.5 & 83.3 & \textcolor{blue}{\textbf{87.2}} \\ 
		\Xhline{1pt}
		\multicolumn{1}{c|}{Res50-Pre-Ours~(736)}             & 92.1 & 83.8 & \textcolor{red}{\textbf{87.8}}  \\ 
		\Xhline{1pt}
	\end{tabular}
	\label{table4}
\end{table}

\begin{figure*}
	\centering
	\includegraphics[width=.9\textwidth]{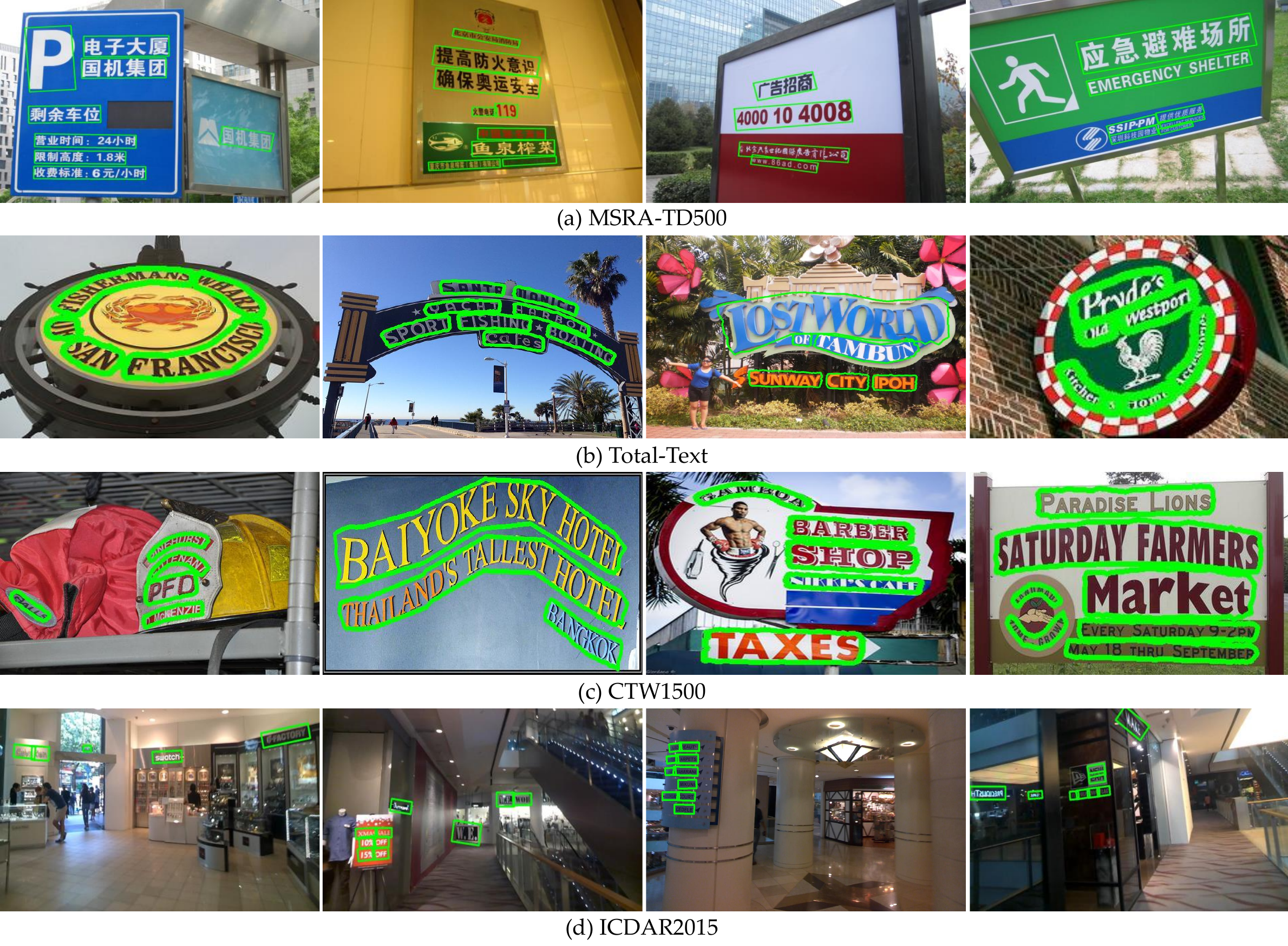}
	\caption{Visualization of the differences between the text contour reconstruction processes with MOS and w/o MOS. The sample is picked from MSRA-TD500 dataset and the $\rm N_d$ and $\rm N_p$ of model are set to 8 and 5, respectively.}
	\label{V11}
\end{figure*}

\textbf{Evaluation on MSRA-TD500.} To verify the performance for detecting line-level multi-oriented text instances, we evaluate the proposed LeafText on the MSRA-TD500 dataset. As shown in Table~\ref{table4}, for existing state-of-the-art (SOTA) methods, ReLaText~\cite{ma2021relatext}, GV~\cite{xu2020gliding}, and DC~\cite{cai2022arbitrarily} achieve 86.7\%, 86.5\%, and 85.4\% in F-measure. Benefiting from decomposing long texts into multiple characters and the strong connection ability of Graph Convolutional Network (GCN), ReLaText surpasses GV and DC in F-measure 0.2\% and 1.3\% respectively. Unlike ReLaText, LeafText models the whole text directly, which effectively avoids the character ignorance problem and improves detection performance. Specifically, our method achieves 87.8\% in F-measure on MSRA-TD500, which surpasses the best existing method ReLaText 1.1\%. For DB++~\cite{liao2022real}, though it achieves significant improvement by embedding DConv~\cite{DBLP:conf/iccv/DaiQXLZHW17} into the corresponding backbone, our method still outperforms it with basic network. We show some qualitative results on MSRA-TD500 in Fig.~\ref{V11}~(a). The above results demonstrate the superior ability of LeafText for detecting very long, multi-oriented, and multi-lingual texts.

\begin{table*}[]
	\renewcommand{\arraystretch}{1.3}
	\setlength{\tabcolsep}{3.5mm}
	\caption{Performance comparison on Total-Text and CTW1500 datasets.}
	\centering
	\begin{tabular}{c|ccc|ccc}
		\Xhline{1pt}
		
		\multirow{2}{*}{Methods}   & \multicolumn{3}{c|}{Total-Text}                                           & \multicolumn{3}{c}{CTW1500}                                             \\ \cline{2-7} 
		&\multicolumn{1}{c}{Precision} & \multicolumn{1}{c}{Recall} & F-measure & \multicolumn{1}{c}{Precision} & \multicolumn{1}{c}{Recall} & F-measure \\ \Xhline{1pt}
		\multicolumn{1}{r|}{TextSnake~\cite{long2018textsnake} (ECCV 2018)}& \multicolumn{1}{c}{82.7}          & \multicolumn{1}{c}{74.5}       &   78.4        & \multicolumn{1}{c}{67.9}          & \multicolumn{1}{c}{85.3}       &   75.6        \\ 
		
		\multicolumn{1}{r|}{ATTR~\cite{wang2019arbitrary} (CVPR 2019)}     & \multicolumn{1}{c}{80.9}          & \multicolumn{1}{c}{76.2}       &   78.5        & \multicolumn{1}{c}{80.1}          & \multicolumn{1}{c}{80.2}       &   80.1        \\ 
		
		\multicolumn{1}{r|}{CRAFT~\cite{baek2019character} (CVPR 2019)    }& \multicolumn{1}{c}{87.6}          & \multicolumn{1}{c}{79.9}       &   83.6        & \multicolumn{1}{c}{86.0}          & \multicolumn{1}{c}{81.1}       &   83.5        \\ 
		
		\multicolumn{1}{r|}{CTD~\cite{qin2019curved} (ICDAR 2019)}         & \multicolumn{1}{c}{80.6}          & \multicolumn{1}{c}{82.3}       &   81.4        & \multicolumn{1}{c}{79.9}          & \multicolumn{1}{c}{77.0}       &   78.5        \\ 
		
		\multicolumn{1}{r|}{LOMO~\cite{zhang2019look} (CVPR 2019)         }&      \multicolumn{1}{c}{87.6}     & \multicolumn{1}{c}{79.3}       &   83.3        & \multicolumn{1}{c}{85.7}          & \multicolumn{1}{c}{76.5}       &   80.8        \\ 
		
		\multicolumn{1}{r|}{PSE~\cite{wang2019shape} (CVPR 2019)}          &      \multicolumn{1}{c}{84.0}     & \multicolumn{1}{c}{78.0}       &   80.9        & \multicolumn{1}{c}{84.8}          & \multicolumn{1}{c}{79.7}       &   82.2        \\ 
		
		\multicolumn{1}{r|}{SegLink++~\cite{tang2019seglink++} (PR 2019)}&  \multicolumn{1}{c}{82.1}           & \multicolumn{1}{c}{80.9}       &   81.5        & \multicolumn{1}{c}{82.8}          & \multicolumn{1}{c}{79.8}       &   81.3        \\ 
		
		\multicolumn{1}{r|}{TextDragon~\cite{feng2019textdragon} (ICCV 2019)}& \multicolumn{1}{c}{85.6}        & \multicolumn{1}{c}{75.7}       &   80.3        & \multicolumn{1}{c}{84.5}          & \multicolumn{1}{c}{82.8}       &   83.6        \\ 
	
		\multicolumn{1}{r|}{Boundary~\cite{wang2020all} (AAAI 2020)}       &   \multicolumn{1}{c}{85.2}        & \multicolumn{1}{c}{83.5}       &   84.3        & \multicolumn{1}{c}{--}            & \multicolumn{1}{c}{--}         &       --   \\ 
		
		\multicolumn{1}{r|}{ContourNet~\cite{wang2020contournet} (CVPR 2020)}& \multicolumn{1}{c}{86.9}        & \multicolumn{1}{c}{83.9}       &   85.4        & \multicolumn{1}{c}{83.7}          & \multicolumn{1}{c}{84.1}       &   83.9        \\ 
		
		\multicolumn{1}{r|}{TextRay~\cite{wang2020textray} (ACMMM 2020)}   & \multicolumn{1}{c}{83.5}          & \multicolumn{1}{c}{77.9}       &   80.6        & \multicolumn{1}{c}{82.8}          & \multicolumn{1}{c}{80.4}       &   81.6        \\ 
		
		\multicolumn{1}{r|}{Spotter~\cite{8812908} (TPAMI~2021)}           &     \multicolumn{1}{c}{88.3}      & \multicolumn{1}{c}{82.4}       &   85.2        & \multicolumn{1}{c}{--}            & \multicolumn{1}{c}{--}         &   --        \\ 
		
		\multicolumn{1}{r|}{FCENet~\cite{zhu2021fourier} (CVPR 2021)}      &      \multicolumn{1}{c}{87.4}     & \multicolumn{1}{c}{79.8}       &   83.4        & \multicolumn{1}{c}{85.4}          & \multicolumn{1}{c}{80.7}       &   83.1        \\ 
		
		\multicolumn{1}{r|}{PSE+STKM~\cite{wan2021self} (CVPR 2021)}       &    \multicolumn{1}{c}{86.3}       & \multicolumn{1}{c}{78.4}       &   82.2        & \multicolumn{1}{c}{85.1}          & \multicolumn{1}{c}{78.2}       &   81.5        \\ 
		
		\multicolumn{1}{r|}{OPMP~\cite{zhang2020opmp} (TMM 2021)}          &       \multicolumn{1}{c}{88.5}    & \multicolumn{1}{c}{82.9}       &   85.6        & \multicolumn{1}{c}{85.1}          & \multicolumn{1}{c}{80.8}       &   82.9        \\ 

		\multicolumn{1}{r|}{ASTD~\cite{dai2021accurate} (TMM 2022)}        &         \multicolumn{1}{c}{85.4}  & \multicolumn{1}{c}{81.2}       &   83.2        & \multicolumn{1}{c}{86.2}          & \multicolumn{1}{c}{80.4}       &   83.2        \\ 
		
		\multicolumn{1}{r|}{TextDCT~\cite{su2022textdct} (TMM 2022)}       &         \multicolumn{1}{c}{87.2}  & \multicolumn{1}{c}{82.7}       &   84.9        & \multicolumn{1}{c}{85.0}          & \multicolumn{1}{c}{85.3}       &   \textcolor{green}{\textbf{85.1}}        \\ 
		
		\multicolumn{1}{r|}{LPAP~\cite{fu2022learning} (TOMM 2022)}        &         \multicolumn{1}{c}{87.3}  & \multicolumn{1}{c}{79.8}       &   83.4        & \multicolumn{1}{c}{84.6}          & \multicolumn{1}{c}{80.3}       &   82.4        \\ 
		
		\multicolumn{1}{r|}{DC~\cite{cai2022arbitrarily} (PR 2022)}        &         \multicolumn{1}{c}{90.5}  & \multicolumn{1}{c}{82.7}       &   \textcolor{blue}{\textbf{86.4}}& \multicolumn{1}{c}{86.9}          & \multicolumn{1}{c}{82.7}       &   84.7        \\ 
		
		\multicolumn{1}{r|}{Res50-DB++~\cite{liao2022real} (TPAMI 2022)}   &         \multicolumn{1}{c}{88.9}  & \multicolumn{1}{c}{83.2}       &   \textcolor{green}{\textbf{86.0}}& \multicolumn{1}{c}{87.9}          & \multicolumn{1}{c}{82.8}       &   \textcolor{blue}{\textbf{85.3}}        \\ 
		
		\Xhline{1pt}
		\multicolumn{1}{c|}{Res18-Pre-Ours~(640)}                          &          \multicolumn{1}{c}{90.8} & \multicolumn{1}{c}{84.0}       &   \textcolor{red}{\textbf{87.3}} & \multicolumn{1}{c}{87.1}          & \multicolumn{1}{c}{83.9}       &  \textcolor{red}{\textbf{85.5}}        \\ \Xhline{1pt}
	\end{tabular}
	\label{table5}
\end{table*}

\textbf{Evaluation on Total-Text and CTW1500.} Irregular-shaped texts bring challenges to existing text detection methods. LeafText represents contours through point sequences for improving the text fitting ability. To verify the effectiveness of our method for the detection of irregular-shaped texts, we make comparisons on the Total-Text and CTW1500 simultaneously. We first resize the short sizes of images into 640 while keeping the original ratio and evaluate the model performance with the backbones of ResNet-18 and ResNet-50, respectively. 

As we can see from Table~\ref{table5}, for the detection of word-level text instances in Total-Text, DC~\cite{cai2022arbitrarily} and DB++~\cite{liao2022real} achieve 86.4\% and 86.0\% in F-measure, they can surpass previous methods up to 7.5\%. On this challenging dataset, LeafText achieves the SOTA performance of 87.3\% in F-measure and exceeds DC~\cite{cai2022arbitrarily} by 0.9\%, which demonstrates the effectiveness of the proposed LVT and the superiority over the existing text representation methods. Meanwhile, the thin vein is helpful for detecting large-scaled instances, which further improves the model detection performance on the Total-Text.

Different from Total-Text, CTW1500 is composed of line-level text instances that contain large spaces between different characters or words, which brings challenges to existing methods. As shown in Table~\ref{table5}, DB++~\cite{liao2022real} and TextDCT~\cite{su2022textdct} are latest SOTA methods on CTW1500 benchmark. They achieve 85.3\% and 85.1\% in F-measure, respectively. A similar conclusion on the CTW1500 dataset can be generated that our method is superior to previous methods. Specifically, our method achieves 85.5\% in F-measure, which surpasses DB++~\cite{liao2022real} 0.2\% even it is equipped with DConv~\cite{DBLP:conf/iccv/DaiQXLZHW17} and complicated backbone (ResNet-50). The experimental results on Total-Text and CTW1500 prove the superiority of the proposed LVT for fitting irregular-shaped texts. Meanwhile, the strong ability to effectively detect word-level and line-level instances simultaneously is verified. Some qualitative results on Total-Text and CTW1500 are depicted in Fig.~\ref{V11}~(b) and (c) for further demonstrating the effectiveness of LeafText. 

\begin{table}[]
	\renewcommand{\arraystretch}{1.3}
	\setlength{\tabcolsep}{1.8mm}
	\caption{Performance comparison on ICDAR2015 Dataset.}
	\centering
	\begin{tabular}{c|ccc}
		\Xhline{1pt}
		Methods                                               &  Precision  &   Recall  &   F-measure \\ \Xhline{1pt}
		WordSup~\cite{hu2017wordsup} (ICCV 2017)              & 79.3 & 77.0 & 78.2  \\ 
		MCN~\cite{DBLP:conf/cvpr/LiuLYFLG18} (CVPR~2018)      & 72.0 & 80.0 & 76.0  \\
		PixelLink~\cite{DBLP:conf/aaai/DengLLC18} (AAAI~2018) & 85.5 & 82.0 & 83.7  \\
		TextBoxes++~\cite{liao2018textboxes++} (TIP~2018)     & 87.8 & 78.5 & 82.9  \\
		PSE~\cite{wang2019shape} (CVPR~2019)                  & 86.9 & 84.5 & 85.7  \\
		RRD~\cite{liao2018rotation} (CVPR~2019)               & 88.0 & 80.0 & 83.8  \\
		SegLink++~\cite{tang2019seglink++} (PR~2019)          & 83.7 & 80.3 & 82.0  \\
		Boundary~\cite{wang2020all} (AAAI~2020)               & 88.1 & 82.2 & 85.0  \\
		FCENet~\cite{zhu2021fourier} (CVPR~2021)              & 85.1 & 84.2 & 84.6  \\
		Spotter~\cite{8812908} (TPAMI~2021)                   & 85.8 & 81.2 & 83.4  \\
		PAN++~\cite{wang2021pan++} (TPAMI~2021)               & 85.9 & 80.4 & 83.1  \\
		EAST+STKM~\cite{wan2021self} (CVPR~2021)              & 88.7 & 84.9 & \textcolor{red}{\textbf{86.8}}   \\
		PSE+STKM~\cite{wan2021self} (CVPR~2021)               & 87.8 & 84.1 & 85.9  \\
		ASTD~\cite{dai2021accurate} (TMM~2022)                & 87.2 & 81.3 & 84.1  \\
		TextDCT~\cite{su2022textdct} (TMM~2022)               & 88.9 & 84.8 & \textcolor{red}{\textbf{86.8}}   \\
		LPAP~\cite{fu2022learning} (TOMM~2022)                & 88.7 & 84.4 & \textcolor{blue}{\textbf{86.5}}  \\
		\Xhline{1pt}
		Res50-Pre-Ours~(1152)                                 & 88.9 & 82.3 & \textcolor{green}{\textbf{86.1}} \\ 	
		\Xhline{1pt}
	\end{tabular}
	\label{table6}
\end{table}

\begin{table}[]
	\renewcommand{\arraystretch}{1.3}
	\setlength{\tabcolsep}{1.2mm}
	\caption{Cross-dataset evaluations on word-level (ICDAR2015 and Total-Text) and line-level (MSRA-TD500 and CTW1500) datasets.}
	\centering
	\begin{tabular}{c|c|c|c|ccc}
		\Xhline{1pt}
		Type                        & Methods   & Training            & Testing            & P & R & F \\ 
		\Xhline{1pt}
		\multirow{6}{*}{word-level}&TextField~\cite{xu2019textfield}&\multirow{3}{*}{IC15}&\multirow{3}{*}{TT}&61.5&65.2&63.3\\ 
		& CM-Net~\cite{DBLP:journals/tip/YangCXYW22}&&&75.8&64.5&69.7\\
		&Res18-Pre-Ours&&&89.2&80.0&84.4\\\cline{2-7}
		&TextField~\cite{xu2019textfield}&\multirow{3}{*}{TT}&\multirow{3}{*}{IC15}&77.1&66.0&71.1\\
		&CM-Net~\cite{DBLP:journals/tip/YangCXYW22}&&&76.5&68.1&72.1\\
		&Res18-Pre-Ours&&&83.0&69.9&75.9\\
		\Xhline{1pt}
		
		\multirow{6}{*}{line-level}&TextField~\cite{xu2019textfield}&\multirow{3}{*}{MSRA}&\multirow{3}{*}{CTW}&75.3&70.0&72.6\\
		&CM-Net~\cite{DBLP:journals/tip/YangCXYW22}&&&77.2&69.7&72.8\\
		&Res18-Pre-Ours&&&83.8&75.0&79.2\\\cline{2-7}
		&TextField~\cite{xu2019textfield}&\multirow{3}{*}{CTW}&\multirow{3}{*}{MSRA}&85.3&75.8&80.3\\
		&CM-Net~\cite{DBLP:journals/tip/YangCXYW22}&&&85.8&77.1&81.2\\
		&Res18-Pre-Ours&&&82.9&82.0&82.4\\ 
		\Xhline{1pt}
	\end{tabular}
	\label{table7}
\end{table}

\textbf{Evaluation on ICDAR2015.} The images in International Conference on Document Analysis and Recognition (ICDAR) 2015 are sampled from the market, which leads to complicated backgrounds and brings challenges to distinguishing texts from interference regions. Moreover, multi-oriented and multi-scaled instance shapes aggravate the difficulty of text detection. To testify the model performance under a complex environment, we conduct comparison experiments on the ICDAR2015 benchmark. As exhibited in Table~\ref{table6}, our method achieves 86.1\% F-measure. Although LeafText is a little lower (0.7\% and 0.4\%) than TextDCT~\cite{su2022textdct} and LPAP~\cite{fu2022learning} in F-measure, our method exceeds most existing SOTA methods (such as PSE~\cite{wang2019shape}, Boundary~\cite{wang2020all} and ASTD~\cite{dai2021accurate}). It is mainly because of LeafText's strong ability to fit various instance shapes and recognize text features. The results in Table~\ref{table6} and Fig.~\ref{V11}~(d) demonstrate our method can recognize the texts with various scales and multi-orientations from the complex background effectively.

\subsection{Cross Dataset Text Detection}
\label{Cross}
To testify the LeafText's generalization performance on different datasets, we evaluate it through cross-train-test experiments. Specifically, the above four public benchmarks are composed of word-level (Total-Text and ICDAR2015) and line-level (MASRA-TD500 and CTW1500) texts. We conduct cross-train-test experiments on the two types of benchmarks in this section, respectively. As shown in Table~\ref{table7}, on the word-level datasets, our method achieves 84.4\% and 75.9\% in F-measure when it is trained on ICDAR2015 and Total-Text and is tested on each other. For line-level datasets, LeafText achieves 79.2\% and 82.4\% in F-measure when it is trained on MSRA-TD500 and CTW1500. The experiments show the LeafText's superior generalization performance.

\begin{figure}
	\centering
	\subfigure[False-positive sample]{
		\begin{minipage}[b]{0.47\linewidth}
			\includegraphics[width=1\linewidth]{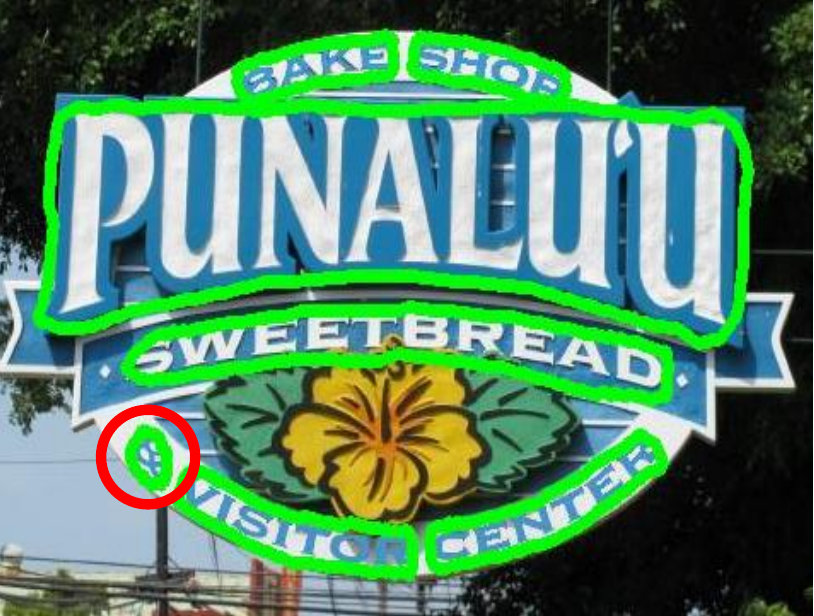}
	\end{minipage}}
	\subfigure[Over emitting]{
		\begin{minipage}[b]{0.47\linewidth}
			\includegraphics[width=1\linewidth]{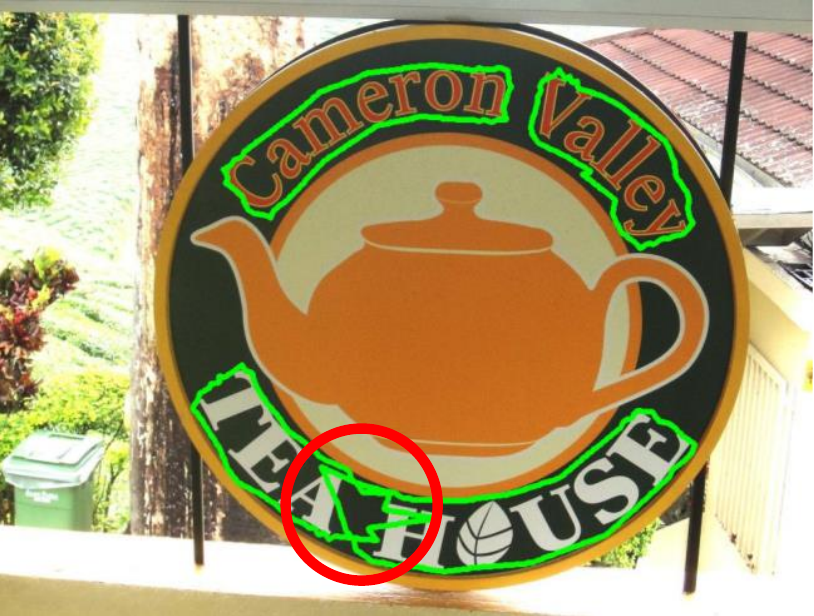}
	\end{minipage}}
	\caption{Illustration of some challenging samples. The green bounding boxes are the detection results from our method. The red ones are failed detection regions.}
	\label{V12}
\end{figure}

\subsection{Limitations of Our Algorithm}
We have analyzed the upper bound performance of LeafText for fitting arbitrary-shaped text instances and verified the effectiveness of LVT, MOS, and thin vein by the ablation studies in Section~\ref{Ablation Study}. Meanwhile, the superior detection and generalization performance on multiple benchmarks of our method are demonstrated in Section~\ref{Comparison} and Section~\ref{Cross}. In this section, we discuss the limitations of our method by visualizing some difficult samples. As depicted in Fig.~\ref{V12}, there are two typical cases. For the false-positive sample (Fig.~\ref{V12}(a)), the high similar vision features between texts and interference regions make it hard to distinguish them effectively. For the case shown in Fig~\ref{V12}(b), there are two adjacent texts and our method over emits into the inner of each other, which brings interference information into detection results and influences the following text recognition task. Therefore, solving the aforementioned limitations that exist in our method will be our future work.

\section{Conclusion}
\label{Conclusion}
In this paper, we explore the leaf vein geometric characteristic and relate it to text contour for designing an effective text representation method (LVT), which improves text fitting ability and avoids disordered point sequence problems naturally. Meanwhile, LVT fining text contour through the thin vein that enjoys half the length of the lateral vein, which reduces the model's complexity and eases the training convergence process while ensuring superior detection performance. Furthermore, considering the lateral and thin veins that are responsible for sampling contour point sequence deeply depending on the main vein, Multi-Oriented Smoother (MOS) enhances the robustness of the main vein, which ensures the correct growth directions of lateral and thin veins effectively. In the end, we successfully accelerate the supervision of lateral and thin vein predictions and balance the importance of texts with different scales through the proposed global incentive loss. Extensive experiments verify the effectiveness of the proposed LVT, MOS, and global incentive loss, and the superiority of the thin vein. Comparisons on the multiple public benchmarks demonstrate the superior detection performance of our approach.

\bibliographystyle{IEEEtran}
\bibliography{egbib}

\vspace{11pt}


\vfill

\end{document}